\documentclass[lettersize,journal]{IEEEtran}
\usepackage{amsmath,amsfonts}
\usepackage{algorithmic}
\usepackage{algorithm}
\usepackage{array}
\usepackage[caption=false,font=normalsize,labelfont=sf,textfont=sf]{subfig}
\usepackage{textcomp}
\usepackage{stfloats}
\usepackage{url}
\usepackage{verbatim}
\usepackage{graphicx}
\usepackage{cite}
\usepackage{color}
\usepackage{multirow}
\usepackage{acronym}
\usepackage{tabularx}
\usepackage{threeparttable} 
\usepackage{makecell}
\usepackage{booktabs}
\usepackage{orcidlink}
\hypersetup{hidelinks}
\usepackage{pifont}
\usepackage{hyperref}
\usepackage{xcolor}
\newcommand{\revise}[1]{\textcolor{black}{#1}}

\begin{document}

\title{ATT-CR: Adaptive Triangular Transformer for Cloud Removal}

\author{Yang Wu, Ye Deng,~\IEEEmembership{Member,~IEEE,} Pengna Li, Wenli Huang, \\ Kangyi Wu, Xiaomeng Xin, and Jinjun Wang
\thanks{This work was supported in part by the National Natural Science Foundation of China under Grants 62088102, U24A20325, and 12326608; the Fundamental Research Funds for the Central Universities under Grant JBK2103012, the Sichuan Science Foundation Project under Grant 2024ZDZX0002 and Grant 2024NSFTD0054, and in part by the Public Welfare Research Program of Ningbo City under Grant 2024S063. (Corresponding author: Ye Deng.)}
\thanks{Yang Wu, Pengna Li, Xiaomeng Xin, Kangyi Wu, and Jinjun Wang are with the Xi'an Jiaotong University, Xi'an, Shaanxi 710049, China (e-mail: wuyang\_cc@stu.xjtu.edu.cn).
    
Ye Deng is with the School of Computing and Artificial Intelligence, Southwestern University of Finance and Economics, Chengdu, Sichuan 611130, China (e-mail: dengye@swufe.edu.cn).
    
Wenli Huang is with the Ningbo University of Technology, Ningbo, Zhejiang 315211, China.}
}

\markboth{Journal of \LaTeX\ Class Files,~Vol.~x, No.~x, x~2025}%
{Shell \MakeLowercase{\textit{et al.}}: A Sample Article Using IEEEtran.cls for IEEE Journals}

\IEEEpubid{0000--0000/00\$00.00~\copyright~2021 IEEE}

\maketitle

\begin{abstract}

\revise{Cloud removal aims to accurately reconstruct the ground objects obscured by clouds in remote sensing images. Existing Transformer-based methods utilizing self-attention have shown impressive results by effectively modeling long-range dependencies in cloudy images}. However, they suffer from the following issues: 1)~the high computational complexity of self-attention limits scalability; 2)~treating both cloudy and clean pixels as valid within the attention computation brings disturbances in subsequent layers, leading to suboptimal performance. To address these challenges, we propose the Adaptive Triangular Transformer for Cloud Removal (ATT-CR), a model that effectively reduces computational costs and mitigates interference from cloudy pixels. Specifically, it consists of two core components: Triangular Attention (TAN) and Feature Selected Gating Module (FSGM). TAN employs lower and upper triangular matrices to approximate Softmax attention with $\mathcal{O}(N)$ computational complexity, significantly reducing the computational costs. The FSGM, on the other hand, integrates with TAN to adaptively distinguish between cloudy and clean features, which minimizes the introduction of invalid information into subsequent layers. Extensive experiments on cloud removal benchmarks demonstrate that ATT-CR delivers superior performance compared to existing methods.

\end{abstract}


\section{Introduction}

\IEEEPARstart{C}loud contamination \revise{significantly impairs the usability of optical remote sensing images in national defense and geoscience applications.} Effectively removing clouds and recovering underlying ground information has emerged as a critical research area in remote sensing image processing. To restore high-quality images from cloud-contaminated ones, earlier methods~\cite{tgrs/VermoteTDHM97, liu2014thin, lgrs/LiHA19, tgrs/GuoHL19} rely on physical priors derived from statistical properties of images. While these prior-based methods are interpretable, their performance is limited by complex atmospheric conditions. In contrast, Convolutional Neural Networks (CNNs) are favored in deep learning architectures for their ability to automatically extract features, minimizing the need for extensive preprocessing and domain-specific knowledge. Many CNN-based cloud removal methods~\cite{li2019thin, Thin-Zi-2021, lgrs/YuZP22, zi2023wavelet, Cloud-Guided-Xiang-2024, ding2024robust} effectively learn the mapping between cloudy and cloud-free images. However, clouds vary significantly in density and extent due to variations in atmospheric conditions, often covering large areas in satellite and aerial imagery. The fixed spatial kernels of CNNs primarily emphasize local features, restricting their capacity to model non-local correlations and adapt to diverse cloud formations.

To deal with the diverse cloud formation and effectively model the long-range dependency in occluded ground information, some works~\cite{SPA-gan-2020,huang2024attentive,cvae-DingZX22, chi2023trinity, cmnet, wu2024cr} \revise{have incorporated self-attention mechanisms that fuse information from both local and non-local contexts for cloud removal.} For instance, CMNet~\cite{cmnet}, GLF-CR~\cite{xu2022glf}, and ACA-CRNet~\cite{huang2024attentive} employ the CNN-based local module to capture reliable texture details while utilizing the self-attention to maintain the structure consistency in the recovered images. Although these methods benefit from the strengths of self-attention, offering improved feature representation capabilities, they still face several issues:

\IEEEpubidadjcol 

1) The quadratic computational complexity of self-attention with respect to feature resolution restricts its long-range modeling capabilities in cloud removal tasks. To mitigate this issue, methods such as SPA-GAN~\cite{SPA-gan-2020} and ACA-CRNet~\cite{huang2024attentive} integrate the attention blocks only in middle layers, where feature maps are smaller, but this strategy limits their representation capabilities. CVAE~\cite{cvae-DingZX22} mitigates complexity by splitting the image into large patches to reduce the number of tokens, but this sacrifices fine-grained pixel-level information. CMNet~\cite{cmnet}, Trinity-Net~\cite{chi2023trinity}, and GLF-CR~\cite{xu2022glf} apply self-attention within small windows to lower computational costs, but this requires complex cyclic shifting window operations and have restricted receptive fields. CR-former~\cite{wu2024cr} linearizes the softmax attention to model pixel-level long-range dependency, with $\mathcal{O}(N)$ computational complexity. However, this approach exacerbates the low-rank limitations inherent in the multi-head attention~\cite{BhojanapalliYRR20}, restricting the feature diversity and network expressiveness. Despite efforts to balance long-range dependency modeling and computational efficiency, these methods still compromise feature representation quality. \revise{More recently, Mamba-CR~\cite{Mamba-CR-2025} and CR-Famba~\cite{liu2025cr} applied Mamba~\cite{gu2023mamba}, a state-space model (SSM), to cloud removal. Mamba achieves linear complexity via recursive filtering of 1D token sequences. However, due to the lossy memory mechanism of SSMs, these models are less expressive than attention-based methods~\cite{yu2025mambaout} and often require additional modifications to compensate for information loss.}

2) Existing Transformer-based cloud removal methods \cite{SPA-gan-2020, cvae-DingZX22, xu2022glf, chi2023trinity, cmnet} typically compute attention across the entire feature map without differentiating between clouded and clean regions, which brings disturbances in feature representation and visual artifacts such as color discrepancy and blurriness in restored images. Moreover, since the transmission properties across different optical bands vary in cloud-contaminated scenarios~\cite{lgrs/YuZP22}, the degraded image features often exhibit channel-wise distinctions, where some channels may retain valid information. Consequently, incorporating both spatial and channel-wise properties to discriminate between cloudy and clean features may substantially enhance cloud removal performance.

\begin{figure}[!t]
	\centering
	\includegraphics[width=3.2in]{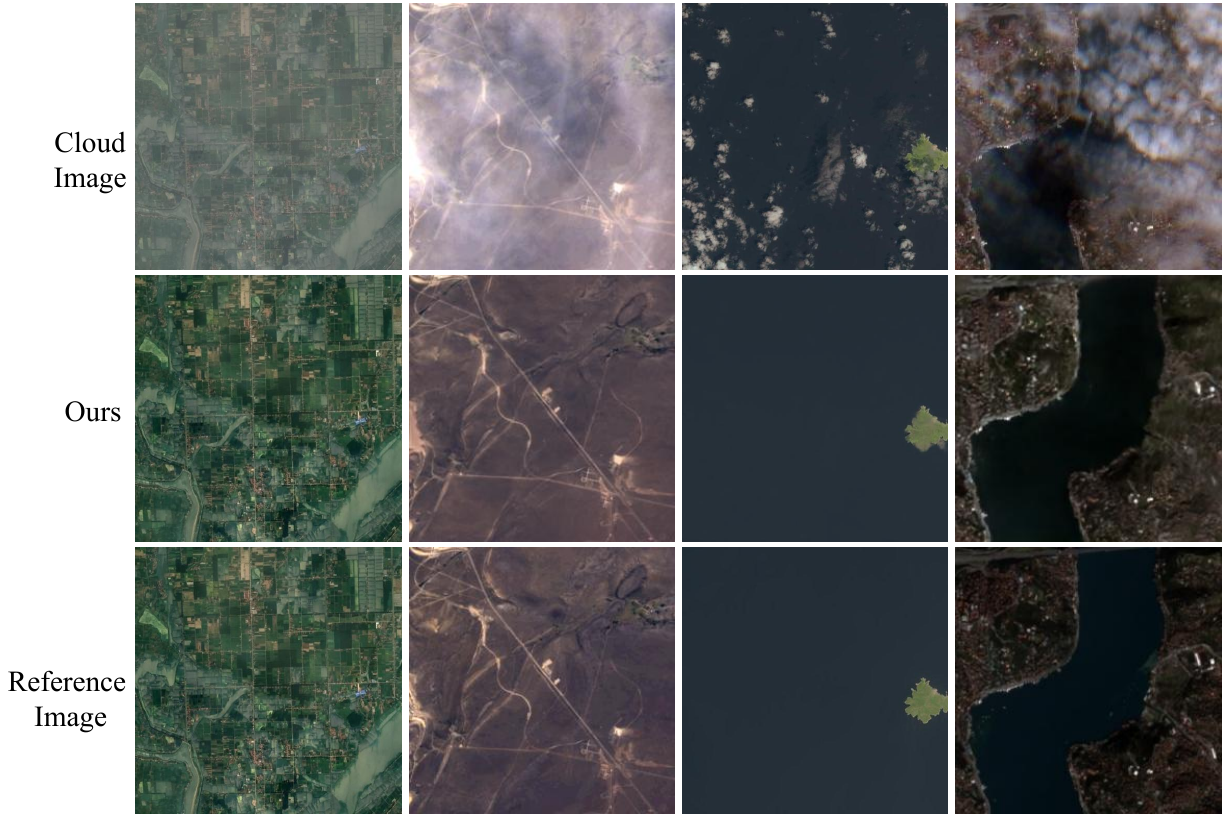}
	\caption{The cloud removal outputs from our ATT-CR model in diverse cloud formations scenarios. The images are sourced from the RICE1, RICE2, T-CLOUD, and SEN12MS-CR datasets. In each column, the first image is the cloudy input, the middle image shows the output from our model, and the last image is the cloud-free reference image.}
	\label{fig:intro}
\end{figure}

To overcome these challenges, we propose the Adaptive Triangular Transformer for Cloud Removal (ATT-CR), which effectively reduces computational costs and mitigates interference from cloudy pixels. Specifically, ATT-CR comprises two core components: Triangular Attention (TAN) and the Feature Selected Gating Module (FSGM). Similar to other methods~\cite{wu2024cr, EA-ShenZZY021, cai2022efficientvit}, TAN first employs a simple activation function to approximate Softmax attention and reorders the attention computation from $(QK^T)V$ to $Q(K^T V)$, achieving a linearized attention mechanism. Subsequently, to address the low-rank limitations of the linear attention~\cite{BhojanapalliYRR20}, our TAN incorporates a triangular attention matrix to preserve the rank of the attention map while maintaining linear computational complexity. This enables the modeling of pixel-level long-range dependencies and high-quality feature representations in cloudy images. To mitigate interference from cloudy pixels in attention computation and consider channel-wise transmission properties, the FSGM adaptively modulates the TAN outputs by differentiating cloudy and clean features at each channel and spatial location. \revise{This process minimizes the propagation of cloudy information into subsequent layers, improving the model's robustness to cloud coverage.} Additionally, we introduce multi-scale tokens into TAN to extract ground object features at varying scales, further enhancing feature diversity and improving overall representation quality.

We perform extensive experiments on real-world datasets, including RICE1, RICE2, T-CLOUD, and the multi-spectral dataset SEN12MS-CR. The experimental results show that ATT-CR delivers superior performance compared to existing methods. Fig.~\ref{fig:intro} shows some restored images of ATT-CR in scenarios with diverse cloud formations. In summary, our main contributions are as follows:

\begin{itemize}
\item We design an ATT-CR network that provides effective feature representation while maintaining computational efficiency for cloudy images.

\item We propose the TAN to capture pixel-level long-range dependencies in cloudy images with $\mathcal{O}(N)$ computational complexity and alleviate the low-rank limitation by employing the multi-head triangular attention matrix.

\item We propose a Feature Selected Gating Module (FSGM) that integrates with TAN to adaptively select features for each channel at every spatial location, enhancing the model’s robustness to cloud coverage.

\end{itemize}

This article follows the structure outlined below. First, we introduce related work on cloud removal in Section~\ref{related works}. Next, section~\ref{methodology} presents our proposed network, ATT-CR, along with its core modules, TAN and FSGM, in a top-down manner. Then, section~\ref{experiments} describes the experimental settings and offers the results along with a detailed analysis. Finally, we summarize the article in section~\ref{conclusion}.

\section{Related Works}
\label{related works}

\subsection{Self-Attention-Enhanced CNNs for Cloud Removal}
CNNs excel at learning complex data distribution within large-scale datasets and show powerful representational ability. Leveraging these advantages, numerous methods~\cite{cvpr/EnomotoSWFMNK17, igarss/SinghK18a, CERMF-Net, Mao-2022} employ CNNs to model the relationship between cloud-free and cloud-contaminated images in an end-to-end manner, directly generating clear images from their cloud-contaminated counterparts. To increase model interpretability, some methods~\cite{LI2020373, lgrs/YuZP22, guo2023blind} integrate the CNNs with physical models, which leverage the neural networks to learn the parameters of traditional physical models. Although these methods have significantly improved cloud removal, the fixed spatial kernels of CNNs limit their capability to model long-range dependencies and adapt to diverse cloud formations.

Thanks to the competence of the self-attention mechanism in modeling long-range dependencies, some works~\cite{SPA-gan-2020, ZhouJ0CSD22, WenPHL22, Chen-Dense-Attention-2023, huang2024attentive} have introduced self-attention to augment CNN-based methods for cloud removal. For example, Pan et al.~\cite{SPA-gan-2020} and Huang et al.~\cite{huang2024attentive} combined convolution with spatial attention to obtain clean images. Zhou et al.~\cite{ZhouJ0CSD22} and Wen et al.~\cite{WenPHL22} employed channel attention mechanisms to suppress clouds and enhance ground details in restored images. Chen et al.~\cite{Chen-Dense-Attention-2023} developed a contextual attention module to match similar clean information from distant regions for cloud removal. However, in CNN-based networks, attention mechanisms are typically confined to deeper layers with smaller feature maps. Although this approach minimizes additional computational overhead, it yields only limited performance improvements.


\subsection{Efficient Transformer for Cloud Removal}
Transformers~\cite{nips/VaswaniSPUJGKP17}, which utilize self-attention to construct a convolution-free neural network, have shown remarkable potential in modeling global information. However, their quadratic computational complexity poses scalability challenges, especially when dealing with the large number of tokens typical in image data. To adapt transformers for high-resolution feature maps, recent works in computer vision have introduced three strategies to reduce computational complexity~\cite{VIT, WangX0FSLL0021, swin-transformer, CSWINDong2022, restormer, wu2025event}: (1) dividing images into large patches to reduce the number of tokens, as in Vision Transformer~\cite{VIT}, (2) constraining attention within small windows and shifting the windows to achieve a global receptive field, as in Swin Transformer~\cite{swin-transformer}, and (3) replacing softmax attention with linear attention, as in Restormer~\cite{restormer}. Building on the success of these strategies, similar approaches have been applied to cloud removal tasks.

CVAE~\cite{cvae-DingZX22} and Trinity-Net~\cite{chi2023trinity} leverage the Vision Transformer~\cite{VIT} and Swin Transformer~\cite{swin-transformer}, respectively, to model degradation factors in cloudy images and restore details from a global perspective. Cloud-EGAN~\cite{ma-Cloud-EGAN-2023} employs the Swin transformer to exploit high-level information from a global perspective.  CMNet~\cite{cmnet} and GLF-CR~\cite{xu2022glf} integrate CNN-based local modules to capture reliable texture details while employing the Swin Transformer~\cite{swin-transformer} to maintain structural consistency in recovered images. Cloudformer~\cite{remotesensing/WuPTH22} strategically applies self-attention within smaller windows to reduce computational demands. While merging tokens and restricting attention to small windows effectively reduce computational complexity, these approaches have notable drawbacks: large patch partitioning sacrifices fine-grained pixel-level information, and shifted window attention limits the receptive field, collectively compromising representation capabilities and reducing the involvement of a wider range of pixels.

To extend attention beyond regional ranges and capture fine-grained information, some researchers reduce the computational burden by replacing Softmax attention with linear attention.  For instance, MDTA~\cite{restormer} applies self-attention across all channels to model pixel-level global information with a computational complexity of $\mathcal{O}(N)$. Building upon MDTA, DFDNet~\cite{Liu-DFDNet-2025} and TSMCF~\cite{Zhu-TSMCF-2025} recover cloudy images to enhance computational efficiency, but it neglects spatial relationships. Another line of research employs simplified activation functions~\cite{linear_transformer, EA-ShenZZY021} or customized mapping functions~\cite{wu2024cr, SOFT} as approximations for the Softmax function in Transformers. These methods exploit the associative property of matrix multiplication to reorder the computation from $(QK^T)V$ to $Q(K^T V)$, achieving $\mathcal{O}(N)$ computation complexity.

Although existing linear attention designs significantly reduce computational costs, they aggravate the low-rank limitation inherent in multi-head attention models~\cite{BhojanapalliYRR20}, which limits their expressiveness. To address this challenge, we develop a simple yet effective attention that utilizes the full-rank property of the triangular matrix combined with the multi-head mechanism, overcoming low-rank limitations and capturing pixel-level long-range dependencies with $\mathcal{O}(N)$ computational complexity.

\subsection{Gated Module}
Gating has been explored in many fields, including computer vision~\cite{hu2018squeeze, yu2019free}, natural language~\cite{dauphin2017language}. For example, Hu et al.~\cite{hu2018squeeze} proposed scaling feature responses by adjusting each channel with learned sigmoid gating values. Yu et al.~\cite{yu2019free} utilized the gating mechanisms to mitigate the influence of invalid pixels in image inpainting. Beyond convolutional gating, Huang et al.~\cite{HUANG2024109897} introduced a sparse self-attention transformer that restricts attention to valid regions, suppressing interference from invalid pixels in inpainting. In cloud removal, remote sensing images are often contaminated by variable and complex cloud cover, where operations on heavily clouded pixels can introduce invalid information, degrading network performance.

To address this issue, Dai et al.~\cite{dai2020gated} incorporated a common convolutional layer with gated convolutional layers to differentiate between cloudy and clean pixels. Wang et al.\cite{wang-CR-2024} integrated a gated convolutional layer in the feature extraction and SAR-optical image fusion modules to mitigate errors caused by cloud regions. Similarly, Wang et al.~\cite{Cloud-Guided-Xiang-2024} employed a Region Gated Module with convolutional operations to distinguish between cloudy and non-cloudy regions.

Cloud-contaminated images exhibit channel-wise characteristics due to varying transmission across color channels. Inspired by gating mechanisms in the aforementioned CNN-based networks and considering channel-wise characteristics of cloud-contaminated image features, we propose a feature selection gating module integrated with our triangular attention module. This design enhances the robustness of TAN against cloud cover by enabling adaptive feature selection across spatial locations and channels, mitigating the impact of cloudy features and improving overall performance.

\section{Methods}
\label{methodology}
We aim to develop a cloud removal network that balances feature representation capability and computational complexity for cloudy remote-sensing images. This section presents the proposed Adaptive Triangular Transformer for Cloud Removal (ATT-CR) in a top-down manner. First, we adopt a multi-stage encoder-decoder architecture as the backbone of ATT-CR. Then, to effectively learn long-range dependencies without constraining the receptive field and alleviate the low-rank limitation inherent in linear attention, we propose the Triangular Attention (TAN) module. Next, to model the varying granularity of ground objects, we integrate Multi-Scale Tokens (MS-Tokens) with TAN to enable the extraction of fine and coarse-grained information. Finally, considering the channel-wise characteristics of degraded image features, we introduce a Feature Selection Gating Module (FSGM) that combines with TAN to adaptively select features for each channel at every spatial location. This enables subsequent layers to focus on the features contributed to cloud removal, enhancing the overall network performance.

\begin{figure*}[!t]
	\centering
	\includegraphics[width=6.4in]{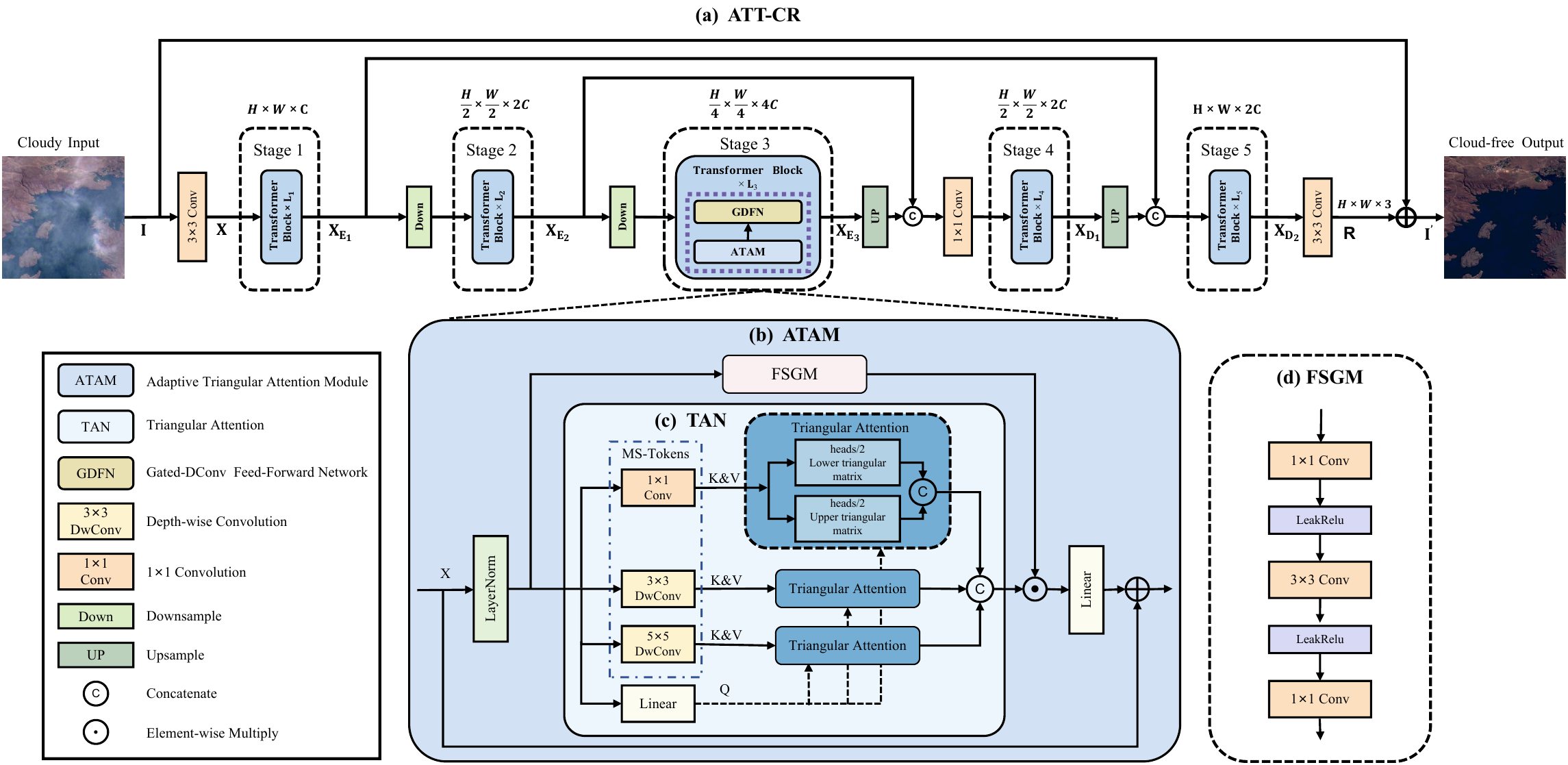}
	\caption{Architecture of ATT-CR. (a) ATT-CR employs a multi-stage design, with each stage consisting of stacked Transformer blocks. (b) The ATAM integrates TAN, FSGM, and Ms-Tokens, working together to achieve high-quality feature representation. (c) TAN has linear computational complexity and addresses the low-rank limitation. (d) FSGM integrates with TAN to minimize the introduction of cloudy features into subsequent layers.}
	\label{fig:MSTAN}
\end{figure*}

\subsection{Overall Pipeline of ATT-CR}
 Our model, ATT-CR, is structured as a multi-stage encoder-decoder architecture, as represented in Fig.~\ref{fig:MSTAN}. Stages 1 to 3 serve as encoders, while stages 4 and 5 are decoders. Each level of the encoder and decoder consists of multiple instances of Transformer blocks. Starting with a cloudy image $I \in \mathbb{R}^{H \times W \times3}$, ATT-CR utilizes a $3 \times 3$ convolution for low-level feature extraction to produce $X \in \mathbb{R}^{H \times W \times C}$. These features are then passed through a five-stage encoder-decoder, followed by a $3 \times 3$ convolutional layer that reduces the channels to 3.
 
\textit{Encoder:} Starting with low-level feature $X \in \mathbb{R}^{H \times W \times C}$, it is processed by three stages (stages 1 to 3) that hierarchically reduce spatial size while increasing channel dimensions. Each stage consists of $L_i$ stacked Transformer blocks and includes a downsampling operation between every two stages. The final encoder stage outputs the feature map is $X_{E_3} \in \mathbb{R}^{\frac{H}{4} \times \frac{W}{4} \times 4C}$.

\textit{Decoder:} The decoder processes the encoder's output \( X_{E_3} \) and progressively restore the spatial resolution. The decoder consists of two stages (Stages 4 to 5). Each stage comprises \( L_i \) stacked Transformer blocks and includes an upsampling operation between consecutive stages. Additionally, a skip connection aggregates information from both the encoder and the decoder, followed by a \(1 \times 1\) convolutional layer to reduce the channel dimension. In the final stage, a \(3 \times 3\) convolutional layer is applied to reduce the channel dimension to 3, generating residual image \( R \in \mathbb{R}^{ H \times W \times 3 } \). The final reconstructed image \( I' \)  is then derived from \( I + R \).

\textit{Transformer Block:} The central Transformer block is composed of an Adaptive Triangular Attention Module (ATAM) and a Gated-DConv Feed-Forward Network (GDFN)~\cite{restormer}. The attention module, ATAM, integrates Multi-Scale Tokens (MS-Tokens) with Triangular Attention (TAN) to model long-range dependencies across multiple scales, while incorporating the Feature Selection Gating Module (FSGM) for feature modulation, thereby enhancing the model's robustness. Following Restormer~\cite{restormer}, GDFN transforms and propagates attention module features to subsequent layers.

\subsection{Triangular Attention}
In the vanilla Transformer~\cite{nips/VaswaniSPUJGKP17}, the input $I_x \in  \mathbb{R}^{N \times d}$ is mapped to query, key, and value through the projection matrices $W_Q$, $W_K$, and $W_V$. Here, $N=H \times W$ represents the resolution of the feature map $I_x$. The self-attention operation is formulated as follows~\cite{nips/VaswaniSPUJGKP17}:
 \begin{equation}
\begin{aligned}
&Q = I_xW_Q, \quad K = I_xW_K, \quad V = I_xW_V, 
\end{aligned}
\label{eq:self-attn-1} 
\end{equation}
\begin{equation}
\begin{aligned}
O_i &= \sum_{j=1}^N  \frac{{\rm sim}(Q_i, K_j) V_j}{\sum_{j=1}^N {{\rm sim}(Q_i, K_j)}},
\end{aligned}
\label{eq:self-attn} 
\end{equation}
where $Q, K, V \in \mathbb{R}^{N \times d}$, $\rm sim(\cdot)$ denotes the similarity function. The current Vision Transformer primarily employs Softmax function, where $\rm{sim}$$(Q, K)=\exp(QK^T)$, to measure similarity. In this case, the term $\exp(QK^T)$ introduces an $\mathcal{O}(N^2)$ computational complexity, resulting in high computational costs for high-resolution features.

To mitigate the computational burden of self-attention, recent works~\cite{linear_transformer, cai2022efficientvit, cosformer, wu2024cr} focus on modifications to the Softmax function, employing carefully designed kernels as approximations of $\exp(QK^T)$. In this way, the similarity function takes the following form:
\begin{equation}
\begin{aligned}
& {\rm sim}(Q, K) =\phi(Q)\phi(K^T).
\end{aligned}
\label{eq:linear_qk} 
\end{equation}
Accordingly, the self-attention formulation in \eqref{eq:self-attn} can be expressed as follows:
\begin{equation}
\begin{aligned}
O_i &=  \frac{\sum_{j=1}^N  \phi(Q_i){\phi(K_j)}^T V_j}{\sum_{j=1}^N \phi(Q_i){\phi(K_j)}^T}.
\end{aligned}
\label{eq:o_i} 
\end{equation}
By applying the associative property of matrix multiplication, the computation in \( (\phi(Q_i) \phi(K_j)^T) V_j \) can be reordered as \( \phi(Q_i) (\phi(K_j)^T V_j) \). 
\begin{equation}
\begin{aligned}
O_i = \phi(Q_i)  \frac{  \sum_{j=1}^N {\phi(K_j)}^T V_j}{\sum_{j=1}^N \phi(Q_i){\phi(K_j)}^T},
\end{aligned}
\label{eq:linear_O} 
\end{equation}  
where the computational complexity becomes \( \mathcal{O}(Nd^2) \). In high-resolution images where the token count reaches tens of thousands \(( N \gg d^2 )\), the computational complexity becomes more favorable. Moreover, the terms \( \sum_{j=1}^N \phi(K_j)^T V_j \) and \( \sum_{j=1}^N \phi(K_j)^T \)  in \eqref{eq:linear_O} are computed only once and shared across all queries. This results in \( \mathcal{O}(N) \) computational and memory complexity, ensuring linear computational cost.

\begin{figure}[!t]
	\centering
	\includegraphics[width=3.0in]{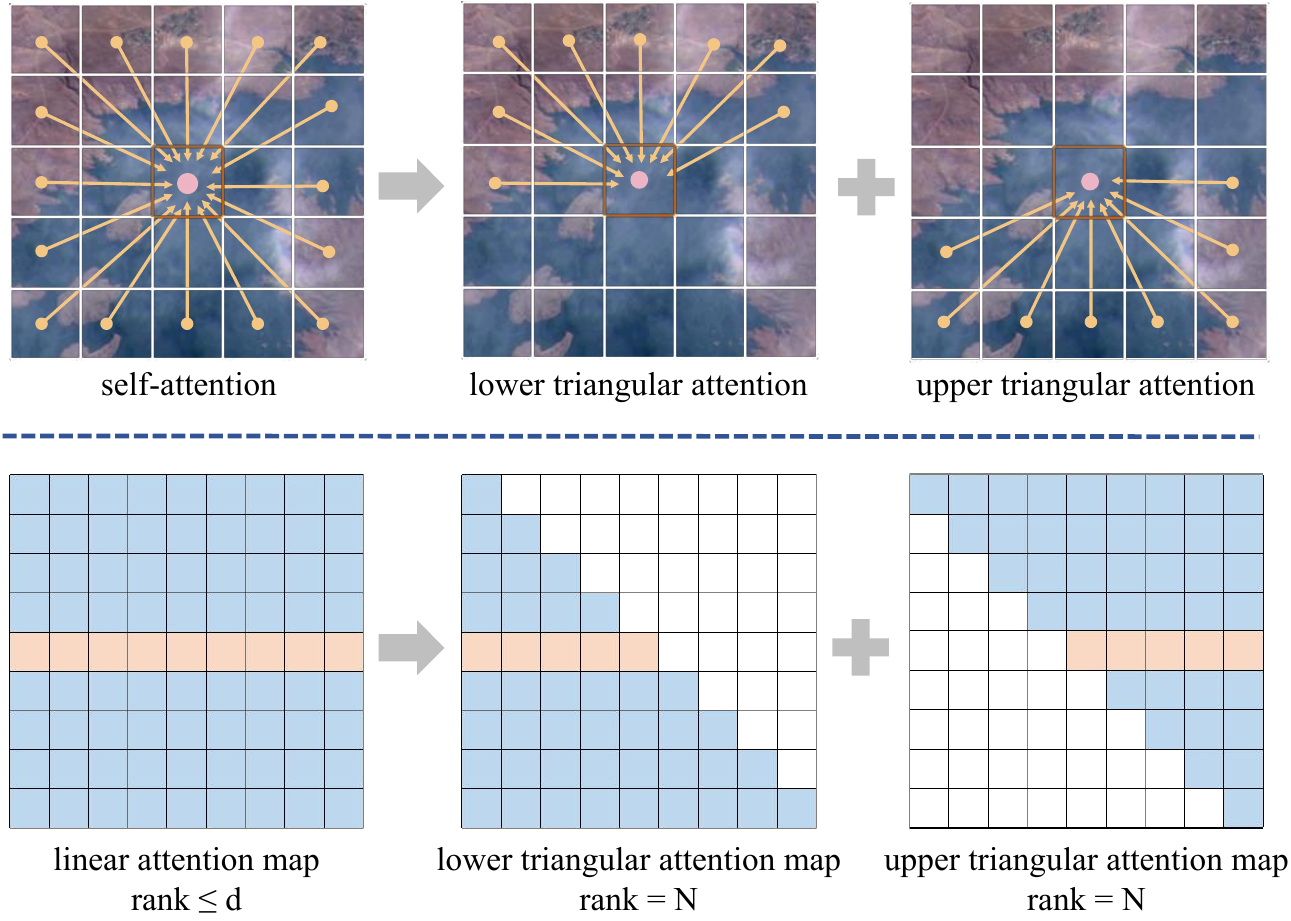}
	\caption{Overview of triangular attention. The first row illustrates the establishment of correlations between image patches, where the orange box represents the query image patch.  The second row displays the attention maps. The original attention is decomposed into two parts: one computes the query with its preceding patches (lower triangular attention) and the other with its subsequent patches (upper triangular attention), achieving a full-rank attention map and global context.} 
	\label{fig:triangular-matrix}
\end{figure}

Although linear attention offers superior computational efficiency compared to Softmax self-attention, it encounters limitations in feature expressiveness. Specifically, the rank of the attention matrix is constrained by the image resolution \( N \) and the channel dimension \( d \)~\cite{BhojanapalliYRR20, flatten-HanPHSH23}. As illustrated below~\cite{flatten-HanPHSH23}:
\begin{equation}
\begin{aligned}
\operatorname{r}(\phi(Q) \phi(K)^T) &\leq \operatorname{min}({\operatorname{r}(\phi(Q)), \operatorname{r}(\phi(K))}) \\
&\leq {\rm min}({N,d}),
\end{aligned}
\label{eq:low-rank} 
\end{equation}
where $r$ denotes the matrix rank. In cloudy remote sensing images $d \ll N$, such as cloud removal inputs with \( d = 48 \) and \( N = 256 \times 256 \). In this scenario, the rank of the attention matrix is constrained by \( d \), resulting in a low-rank property, where multiple rows of the attention map exhibit significant homogenization. Since the output of self-attention is a weighted sum of \( V \), homogenized attention weights render aggregated features indistinguishable, leading to a reduced representation ability.

\begin{figure}[!t]
	\centering
	\includegraphics[width=3.4in]{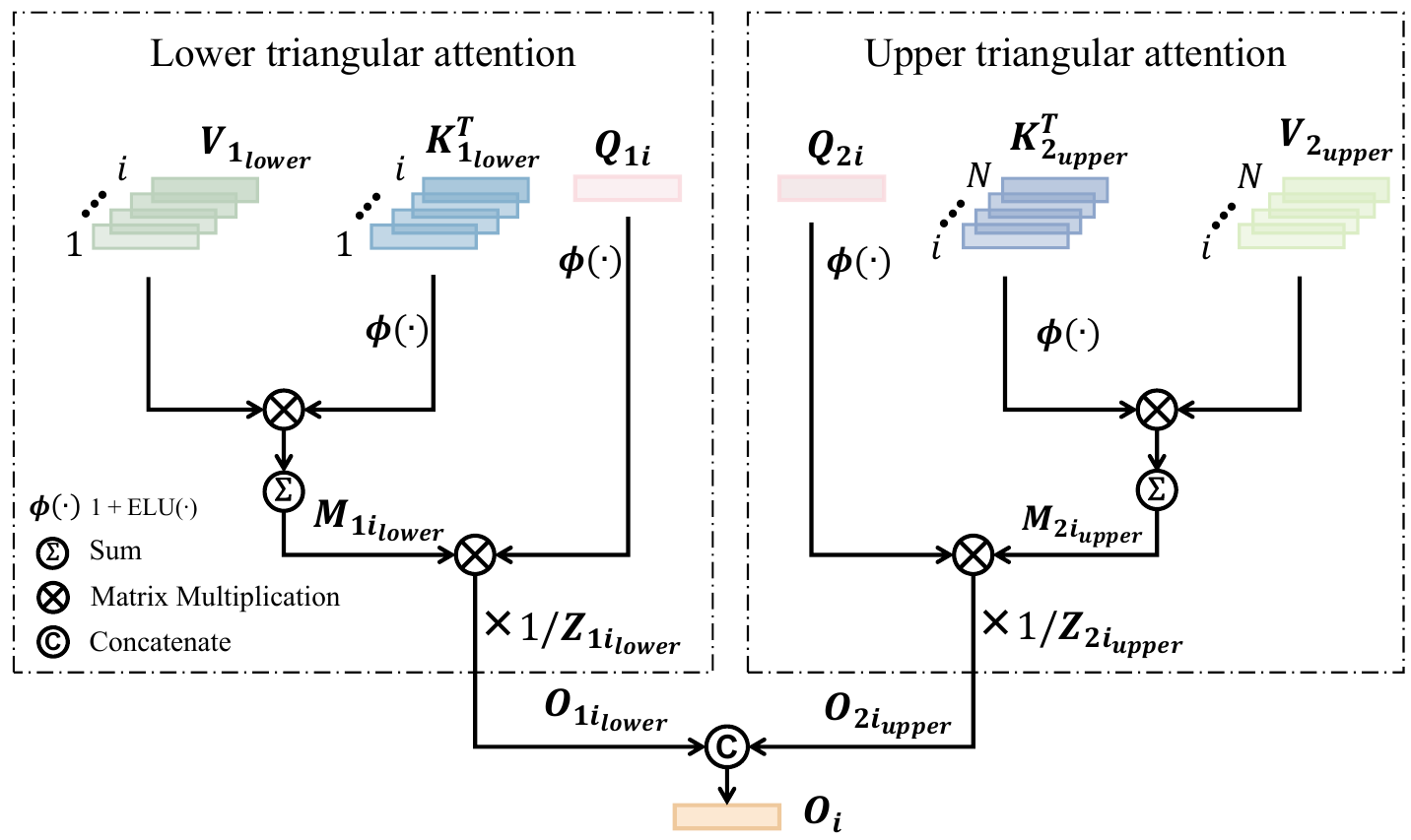}
	\caption{The calculation of the triangular attention output values involves splitting Q, K, and V into multiple heads: half of the heads compute the lower triangular attention output $O_{1i_{lower}}$, while the other half compute the upper triangular attention output $O_{2i_{upper}}$, achieving a full-rank representation while maintaining computational efficiency.}
	\label{fig:triangular-matrix-cal}
\end{figure}

To mitigate this limitation, we design a simple yet effective Triangular Attention mechanism to enhance the rank of the attention map, as presented in Fig.~\ref{fig:triangular-matrix}. Specifically, inspired by the full-rank property of the triangular matrix, we divide the attention heads into two groups: one computes upper triangular attention, while the other computes lower triangular attention.

Firstly, we split Q, K, V into two heads as follows:
\begin{equation}
Q_1, Q_2 = \text{split} (Q), \quad Q\in \mathcal{R}^{N \times d}, Q_1, Q_2 \in \mathcal{R}^{N \times d/2},
\end{equation}
\begin{equation}
K_1, K_2 = \text{split} (K), \quad  K\in \mathcal{R}^{N \times d},  K_1, K_2 \in \mathcal{R}^{N \times d/2},
\end{equation}
\begin{equation}
V_1, V_2=\text{split}(V), \quad  V\in \mathcal{R}^{N \times d},  V_1, V_2 \in \mathcal{R}^{N \times d/2},
\end{equation}

Then, the attention matrices are defined as:
 \begin{equation}
 \mathbf{A}_{lower} = \phi(Q_1) \phi(K_1)^\top \cdot \mathbf{I}_{i \leq j}
    \label{eq:placeholder1}
\end{equation}
\begin{equation}
    \mathbf{A}_{upper} = \phi(Q_2) \phi(K_2)^\top \cdot \mathbf{I}_{i \geq j}
    \label{eq:placeholder2}
\end{equation}

where $\mathbf{I}_{i \leq j}, \mathbf{I}_{i \geq j} \in \mathcal{R}^ {N \times N} $ are indicator functions that apply lower and upper triangular masks, respectively. Here, $i$ is the row index and $j$ is the column index. Specifically, $\mathbf{I}_{i \leq j}$ ensures that each position $i$ can only attend to previous positions (including itself), while $\mathbf{I}_{i \geq j}$ ensures that each position $i$ can only attend to itself and subsequent positions.  Each masked attention matrix is computed independently in separate attention heads.

In this way, the original linear attention is decomposed into two triangular attention branches:
\begin{equation}
 O= \phi(Q)\phi(K)^TV \rightarrow O=\text{Concat}(A_{uper}V_1, A_{lower} V_2)
\label{eq:ORGtoTAN}
\end{equation}
where each attention matrix $\mathbf{A}_{lower}, \mathbf{A}_{upper} \in \mathcal{R}^ {N \times N}$ is a triangular matrix. A well-established property in linear algebra states that: If  $M \in \mathbb{R}^{N \times N}$ is triangular and non-singular, then $\mathrm{rank}(M) = N$.

Therefore, this design ensures that the resulting attention maps $A_{lower}$ and $A_{upper}$ maintain full rank \( N \), effectively mitigating the rank limitation $d$ inherent in conventional linear attention. Additionally, by leveraging the multi-head attention mechanism, our Triangular Attention captures global context by aggregating both upper and lower triangular components in a complementary manner.
 
Since the calculation of triangular attention involves causal masking, its detailed formulation (illustrated in Fig.~\ref{fig:triangular-matrix-cal}) is expressed as follows:
\begin{equation}
\begin{aligned}
&O_{1i_{lower}} = \phi(Q_{1i})  \frac{ \sum_{j=1}^i {\phi(K_{1j})}^T V_{1j}}{\sum_{j=1}^i \phi(Q_{1i}){\phi(K_{1j})}^T}, \\
&O_{2i_{upper}} = \phi(Q_{2i})  \frac{ \sum_{j=i}^N {\phi(K_{2j})}^T V_{2j}}{\sum_{j=i}^N \phi(Q_{2i}){\phi(K_{2j})}^T}.\\
\end{aligned}
\label{eq:tri_attn} 
\end{equation}
We introduce the $M_i$ and $Z_i$ into \eqref{eq:tri_attn}, where
\begin{equation}
\begin{aligned}
 M_{{1i}_{lower}} = \sum_{j=1}^i {\phi(K_{1j})}^T V_{1j}, \quad
M_{{2i}_{upper}} = \sum_{j=i}^N {\phi(K_{2j})}^T V_{2j},
\end{aligned}
\label{eq:M_i} 
\end{equation}
\begin{equation}
\begin{aligned}
Z_{1i_{lower}} = \phi(Q_{1i})\sum_{j=1}^i {\phi(K_{1j})^T}, \\
Z_{2i_{upper}} = \phi(Q_{2i}) \sum_{j=i}^N {\phi(K_{2j})^T}.
\end{aligned}
\label{eq:Z_i} 
\end{equation}
Then, \eqref{eq:tri_attn} is simplify as:
\begin{equation}
\begin{aligned}
O_{{1i}_{lower}} = \frac{ \phi(Q_{1i}) M_{{1i}_{lower}}}{ Z_{{1i}_{lower}}}, \quad
O_{{2i}_{upper}} = \frac{ \phi(Q_{2i}) M_{{2i}_{upper}}}{ Z_{{2i}_{upper}}}.
\end{aligned}
\label{eq:tri_attn_sim} 
\end{equation}
The global information is captured by concatenating the lower triangular attention output \( O_{i_{lower}} \) and the upper triangular attention output \( O_{i_{upper}} \), as seen in Fig.~\ref{fig:triangular-matrix-cal}. The output $O_{i}$ is represented as follows:
\begin{equation}
\begin{aligned}
O_i &= {\rm Concat}(O_{{1i}_{lower}}, O_{{2i}_{upper}}),
\end{aligned}
\label{eq:tri_attn_sim-2} 
\end{equation}
where \( M_i \) and \( Z_i \) are derived from \( M_{i-1} \) and \( Z_{i-1} \) in constant time, resulting in linear computational complexity with respect to the sequence length \( N \). Following the approach in linear attention~\cite{linear_transformer}, \revise{we adopt \( \phi(x) = {\rm ELU}(x) + 1 \)}, where \revise{ELU} ensures positive similarity while preventing the gradients from becoming zero when \( x \) is negative, unlike ReLU.

This simple and effective method utilizes the full-rank property of the triangular matrix combined with the multi-head mechanism, which overcomes low-rank limitations and captures pixel-level long-range dependencies with \( \mathcal{O}(N) \) computational complexity.

Moreover, ground objects in remote sensing images exhibit significant scale variations, from large water bodies to small buildings. \revise{To effectively capture local features at varying granularities and enhance feature diversity, we combine multi-scale tokens with triangular attention, following prior works such as MB-TaylorFormer V2~\cite{jin2025mb}, Gridformer~\cite{wang2024gridformer}, DDMSNet~\cite{zhang2021deep}, and ESTINet~\cite{hsu2023wavelet}.} Specifically, we apply convolution operations with different kernel sizes to the keys and values, generating coarse and fine visual tokens that capture multi-scale information. For computational efficiency, we use small-kernel depthwise separable convolutions~\cite{howard2017mobilenets} to generate multi-scale tokens. The query, key, and value are formulated as follows and the detailed architecture is illustrated in Fig.~\ref{fig:MSTAN}.
 \begin{equation}
\begin{aligned}
&Q = \mathrm{Conv}_{1\times1}(X),\\
&K = \mathrm{DwConv}_{s\times s}({\rm Conv}_{1\times1}(X)), \\
&V = \mathrm{DwConv}_{s\times s}({\rm Conv}_{1\times1}(X)), \\
\end{aligned}
\label{eq:QK-conv} 
\end{equation}
where $s$ is the convolution kernel size. 

\subsection{Feature Selection Gating Module}
Gating mechanisms in convolutions~\cite{yu2019free} have been demonstrated to effectively distinguish valid pixels in features, thereby enhancing the robustness of the network. Similarly, cloud-contaminated images also exhibit the issue of cloud-interfered pixel features. Inspired by this, we introduce a gated mechanism into our proposed triangular attention to differentiate between cloudy and cloud-free pixels, thereby improving the attention's ability to handle cloud interference. Specifically, we propose a Feature Selection Gating Module (FSGM) for cloud removal, which performs element-wise multiplication with the TAN output to modulate it and mitigate the propagation of cloudy features into subsequent layers. Since the transmission properties vary across different optical bands, the features of cloud-contaminated images often exhibit channel-wise distinctions, e.g., some bands can penetrate the clouds and retain certain ground details. The FSGM, considering the channel-wise characteristics adaptively selects features at both channel and spatial levels, dynamically focusing on relevant features at different stages to enhance the model's robustness against cloud contamination.

As illustrated in Fig.~\ref{fig:MSTAN} (d), FSGM employs multiple convolutional layers and LeakyReLU activations to effectively capture spatial and channel-wise features, enabling the network to identify and suppress invalid features caused by cloudy pixels. The gating mechanism is formulated as:
\begin{equation}
\begin{aligned}
G(X) = {\rm Conv_{1\times1}}( {\rm LeakReLu}({\rm Conv_{3\times3}}(X))_{\times 2}),
\end{aligned}
\label{eq:gate} 
\end{equation}
where $\times 2$ denotes two layers combining convolutional operations and LeakyReLU activations.

We integrate FSGM with TAN to guide subsequent network layers. The refined outputs of TAN are obtained by the product of the outputs of FSGM and TAN, which are computed as follows:
\begin{equation}
\begin{aligned}
&\hat{X} = G(X)\odot {\rm Cat}(H_1, H_2, .., H_{i}),\\
& H_i = {\rm TAN}(Q_i, K_i, V_i ),
\end{aligned}
\label{eq:mstan-gate} 
\end{equation}
where $\odot$ represents the Hadamard product and $H_i$ is the output of the $i$-th scale token from TAN. 

G(X) generate dynamical weights based on the input feature $X$ and adaptively modulates the TAN outputs. This allows the model to emphasize clean regions and suppress cloud-contaminated areas, guiding subsequent network layers to focus on informative features. Visualizations of intermediate gating values reveal that FSGM can differentiate between cloudy and cloud-free features in distinct channels. This enables our network to selectively choose valid features in different channels, reducing the interference of cloudy pixels in attention computation and aiding in the restoration of missing information. Further details are provided in Section~\ref{Visualization of Gated values}.

\subsection{Loss Function}
In this article, we optimize the proposed model using the mean squared error (MSE) loss, which aims to minimize the error between the network output and the cloud-free reference image. The loss function is formulated as follows:
\begin{equation}
\begin{aligned}
L = \frac{1}{N} \sum_{i=1}^N||F(X_i,\Theta)-Y_i||^2,
\end{aligned}
\label{eq:loss} 
\end{equation}
where $N$ denotes the number of training images, $F(.)$ represents our network, $X_i$ is the clouded input image, and $Y_i$ is the corresponding clean reference image.

\section{Experiments}
\label{experiments}
To assess the effectiveness of our method, we conduct comprehensive experiments on well-established cloud removal benchmarks and compare the results with several state-of-the-art models. This section details the experimental setup and provides a comprehensive analysis of the results.

\subsection{Experimental Settings}
\subsubsection{Datasets}
We validate our methods on four publicly available cloud removal datasets: RICE1~\cite{RICE1}, RICE2~\cite{RICE1}, T-CLOUD~\cite{cvae-DingZX22}, and SEN12MS-CR.

\textit{RICE1:} The RICE1~\cite{RICE1} dataset from Google Earth consists of 500 image pairs, each containing a clouded and a corresponding clear image. It is a thin-cloud dataset, with all images cropped to a resolution of $512 \times 512$ pixels without overlap. For training and testing, the dataset is separated into 400 and 100 pairs, respectively.

\textit{RICE2:} The RICE2~\cite{RICE1} dataset focuses on thick-cloud scenarios and includes 735 image pairs, each with a resolution of $512 \times 512$ pixels. We split the dataset into 588 pairs for training and 147 pairs for testing.
 
\textit{T-CLOUD:} The T-CLOUD~\cite{cvae-DingZX22} dataset is a real-world thin-cloud dataset captured from Landsat 8 RGB images, consisting of 2,939 image pairs. Cloudy images and their corresponding clear images are captured with a 16-day satellite revisit cycle. They are carefully selected under consistent lighting conditions, with all images cropped to a resolution $256 \times 256$ pixels. For training and testing, the dataset is separated into 2,351 and 588 pairs, respectively.

\textit{SEN12MS-CR:} The SEN12MS-CR~\cite{sen12mscr} is a large-scale multi-spectral dataset for cloud removal. It contains co-registered radar (Sentinel-1) and optical satellite images (Sentinel-2), forming paired samples. Each sample is a triple, including the 2-band Sentinel-1 SAR data and the 13-band Sentinel-2 optical observation data with and without cloud cover. The dataset spans 175 globally distributed regions across four seasons in 2018 and contains a total of 122,218 patches, each with a resolution of $256 \times 256$ pixels. For training and testing, the dataset is separated into 114,325 and 7,893 pairs, respectively.

\subsubsection{Experiment Details} 

Our method is implemented in PyTorch, and the experiments are conducted on a system with four NVIDIA RTX 3090 GPUs running Ubuntu 20.04. The model follows a multi-stage encoder-decoder architecture with Transformer blocks [1, 2, 8, 2, 1], attention heads [2, 2, 8, 2, 2], and channels [48, 96, 192, 96, 96] per stage, following the design~\cite{restormer}. For the T-CLOUD, RICE1, and SEN12MS-CR datasets, the kernel sizes of the multi-scale token convolution are set to ([ [3, 5], [3, 5], [1, 3], [3, 5], [3, 5] ]) across stages and [[3, 5], [1, 3], [1, 3], [1, 3], [3, 5]] for RICE2. The model is trained using the AdamW optimizer with \( L_1 \) loss. The learning rate starts at \( 4 \times 10^{-4} \) and gradually decreases to \( 2 \times 10^{-6} \) through cosine annealing, following the approach in~\cite{wu2024cr, restormer}. Training is performed on \( 256 \times 256 \) patches. For the T-CLOUD, RICE1, and RICE2 datasets, the model is trained for 1,000 epochs with a batch size of 16. For the SEN12MS-CR dataset, we follow~\cite{xu2022glf} and concatenate the 2-channel SAR data with 13-channel multispectral optical images to construct 15-channel inputs. The model is trained for 30 epochs with a batch size of 16 using the MultiStepLR scheduler~\cite{ebel2023uncrtaints}. The learning rate is initially set to \( 6 \times 10^{-4} \) for the first 15 epochs, then halved every 5 epochs for the remaining training.

\subsubsection{Comparison Methods}
To assess our model’s performance, we compared it against eight state-of-the-art (SOTA) cloud removal methods:
\begin{itemize}
	\item{pix2pix~\cite{pix2pix} (2019):} A conditional GAN framework that employs a U-Net generator and a PatchGAN discriminator to learn mappings between paired images.
	\item{SPA-GAN~\cite{SPA-gan-2020} (2020):}  A GAN model with local-to-global spatial attention for identifying cloud regions and generating clean images.

    \item{CVAE~\cite{cvae-DingZX22} (2022):} A conditional variational autoencoder framework that utilizes the Vision Transformer~\cite{VIT} to model the distribution of degradation factors in cloud-contaminated images and reconstruct cloud-free images based on these factors.
   
    \item {Restormer~\cite{restormer} (2022):} A Transformer-based image restoration model that uses channel-wise self-attention with linear complexity to capture pixel-level long-range dependencies efficiently.

    \item{Trinity-Net~\cite{chi2023trinity} (2023):} A novel framework that combines prior information with CNNs and Swin Transformer layers to accurately estimate haze parameters, enabling effective dehazing of remote sensing images.
    \item{CMNet~\cite{cmnet} (2024):} A multistage neural network featuring two complementary subnetworks to refine local spatial details and extract global features using Swin Transformer~\cite{swin-transformer} layers for cloud removal.
    \item{ACA-CRNet~\cite{CTGAN/HuangW22} (2024):} A novel attention-based approach that dynamically selects scores to suppress noise and irrelevant features, enhancing distant context modeling for better cloud removal.
    \item {CR-former~\cite{wu2024cr} (2024):} A U-net-style model leveraging Focused Taylor Attention (FT-Attention), which linearizes softmax attention and produces more distinctive attention weights for efficient feature extraction in cloud removal tasks.
   \item{CR-Famba\cite{liu2025cr} (2025):} A frequency-domain assisted Mamba for cloud removal, which explores the long-range modeling ability of state space models in remote sensing images.
\end{itemize}

\subsection{Experimental Results}
\subsubsection{\textbf{Quantitative Results}}
To assess our method's effectiveness, we perform comprehensive experiments on the real-world datasets RICE1, RICE2, and T-CLOUD. Table~\ref{tab:main-results} provides a comparison between ATT-CR and SOTA models. Following previous studies on cloud removal~\cite{cmnet, xu2022glf, SPA-gan-2020}, we quantitatively evaluate model performance using the following metrics: mean absolute error (MAE)\cite{chai2014root},  spectral angle mapper (SAM)\cite{kruse1993spectral}, peak signal-to-noise ratio (PSNR)\cite{korhonen2012peak}, and structural similarity index (SSIM)\cite{WangBSS04}. Additionally, we report the parameter counts (Params) and floating-point operations (FLOPs), where FLOPs represent the multiply-accumulate operations required for the forward pass, and the parameter count reflects the model's size.

\begin{table*}[tbp]
	\centering
	\caption{ Quantitative comparison on RICE1, RICE2, and T-CLOUD. $\downarrow$: Lower values indicate better performance, $\uparrow$: Higher values indicate better performance, with best results highlighted in bold. } 
    \renewcommand\arraystretch{1.0}
	\setlength{\tabcolsep}{0.5mm}  
        {
		\begin{tabular}{c|cccc|cccc|cccc|cc}
			\toprule
                \multirow{3}[1]{*}{\makecell[c]{\textbf{Models}}}  &\multicolumn{4}{c|}{\textbf{RICE1}} &\multicolumn{4}{c|}{\textbf{RICE2}} &\multicolumn{4}{c|}{\textbf{T-CLOUD}}  & \multicolumn{2}{c} {\textbf{Overhead}}\\
			  &\multirow{2}[1]{*}{\makecell[c] {\textbf{MAE} $\downarrow$}} &\multirow{2}[1]{*}{\makecell[c]{\textbf{SAM} $\downarrow$}} &\multirow{2}[1]{*}{\makecell[c]{\textbf{PSNR}$\uparrow$}}  &\multirow{2}[1]{*}{\makecell[c]{\textbf{SSIM}$\uparrow$}} &\multirow{2}[1]{*}{\makecell[c] {\textbf{MAE} $\downarrow$}}   &\multirow{2}[1]{*}{\makecell[c]{\textbf{SAM} $\downarrow$}} &\multirow{2}[1]{*}{\makecell[c]{\textbf{PSNR}$\uparrow$}}  &\multirow{2}[1]{*}{\makecell[c]{\textbf{SSIM}$\uparrow$}} &\multirow{2}[1]{*}{\makecell[c] {\textbf{MAE} $\downarrow$}}   &\multirow{2}[1]{*}{\makecell[c]{\textbf{SAM} $\downarrow$}} &\multirow{2}[1]{*}{\makecell[c]{\textbf{PSNR}$\uparrow$}}  &\multirow{2}[1]{*}{\makecell[c]{\textbf{SSIM}$\uparrow$}} &\textbf{Param} & \textbf{FLOPs} \\
             &&&& &&&& &&&& &(M)&(G)\\
			\midrule
			pix2pix~\cite{pix2pix}  & 0.0253  & 4.11  & 31.97  & 0.9161  & 0.0327  & 5.08 & 29.45  & 0.8473 & 0.0449  & 10.01  & 25.64  & 0.7563 &54.41 &6.1 \\
			SPA-GAN~\cite{SPA-gan-2020}\tnote{*}  & 0.0372  & 2.25 & 28.89  & 0.9144 & 0.0403  & 3.36  & 27.51  & 0.8177 & 0.0419  & 4.09  & 26.14  & 0.7954 &0.21 &15.2 \\
			CVAE~\cite{cvae-DingZX22}\tnote{*}  & 0.0198  & 1.11  & 33.70  & 0.9562 & 0.0216  & 1.69  & 33.62  & 0.9079 & 0.0342  & 2.95  & 28.19  & 0.8613  &15.42 & 37.1 \\
            Restormer~\cite{restormer} & 0.0176 & 1.01 &35.42 & 0.9617 & 0.0163 & 1.27 & 36.05 & 0.9155 & 0.0248 & 2.46 & 30.49 & 0.8851 & 26.13 &155.0 \\
          
             Trinity-Net~\cite{chi2023trinity}  & 0.0305 &2.20 & 30.12 &0.9605 & 0.0295 & 2.63 &30.09 & 0.8771 &0.0365 & 3.67 & 27.33 & 0.8410 & 20.24 & 17.6   \\
            CMNet~\cite{cmnet} \tnote{*}  & 0.0144  & 0.98  & 36.26  & 0.9625 & 0.0167  & 1.29 & 35.82  & 0.9151 & 0.0251  & 2.49  & 30.33  & 0.8829 &16.51& 236.0\\

            CR-former-L~\cite{wu2024cr}  & ${0.0138} $ & ${0.92}$  & ${36.75}$ & ${0.9627}$ & ${0.0163}$   & ${1.26}$   & ${36.24}$ & ${0.9161}$ & ${0.0238}$   & ${2.43}$   & ${30.82}$   & ${0.8867}$ &26.15  & 155.0 \\
            ACA-CRNet~\cite{huang2024attentive} &0.0151 & 0.98 &36.05 &0.9628&0.0164&1.29&35.65&0.9126&0.0234&2.35&30.85&0.8885&20.39 &1456.1\\
         CR-Famba~\cite{liu2025cr} & 0.0325 & 1.64 & 30.91 & 0.9142 & 0.0252&2.23 & 31.90 & 0.8495 & 0.0262  & 2.48 & 30.10 &  0.8810 & 174.94 & 165.0\\
 
            \midrule
            Ours & $\mathbf{0.0135} $ & $\mathbf{0.91}$  & $\mathbf{36.83}$ &$\mathbf{0.9629}$ & $\mathbf{0.0149}$   & $\mathbf{1.16}$   & $\mathbf{36.72}$ & $\mathbf{0.9183}$ & $\mathbf{0.0226}$   & $\mathbf{2.31}$   & $ \mathbf{31.12}$   & $\mathbf{0.8908}$ &6.93  & 53.3 \\
            
			\bottomrule
		\end{tabular}}
	\label{tab:main-results}%
\end{table*}%

Referring to Table~\ref{tab:main-results}, our model reaches a PSNR value of 36.83dB and obtains an SSIM of 0.9629 on the RICE1 dataset, outperforming the previous SOTA method by 0.08dB in PSNR. On the RICE2 dataset, ours improve PSNR by 0.48dB over the SOTA method. For the T-CLOUD dataset, our model reaches an SSIM of 0.8908 and a PSNR of 31.12dB, with a 0.30dB PSNR gain over the CR-former. These performance gains are primarily attributed to our model overcoming the low-rank limitation inherent in linear attention and the introduction of FSGM which adaptively selects features mitigating the influence of cloudy pixels. In contrast, CR-former does not consider the low-rank limitation and the challenges posed by cloudy pixels, which reduces its effectiveness in handling cloud-contaminated images. For the Mamba-based method CR-Famba, we rerun the publicly available code on our dataset. Compared to CR-Famba, our method achieves significantly better performance while using far fewer parameters (6.93M vs. 174.94M) and FLOPs (53.5G vs. 174.94G), highlighting the efficiency of our method in both performance and computational cost.

\begin{table}[tbp] 
	\centering
	\caption{ Quantitative comparison on the multi-spectral dataset SEN12MS-CR. } 
    \renewcommand\arraystretch{1}
	\setlength{\tabcolsep}{0.5mm}  
        {
		\begin{tabular}{c|cccc |cc}
			\toprule
                \multirow{3}[1]{*}{\makecell[c]{\textbf{Models}}} & \multicolumn{4}{c|}{\textbf{SEN12MS-CR}}  & \multicolumn{2}{c}{\textbf{Overhead}}\\
			 &\multirow{2}[1]{*}{\makecell[c]{\textbf{MAE}$\downarrow$}}
              &\multirow{2}[1]{*}{\makecell[c]{\textbf{SAM}$\downarrow$}}
             &\multirow{2}[1]{*}{\makecell[c]{\textbf{PSNR}$\uparrow$}}  &\multirow{2}[1]{*}{\makecell[c]{\textbf{SSIM}$\uparrow$}}  &\textbf{Param} & \textbf{FLOPs}  \\
             && && &(M)&(G) \\
		
        \midrule
        pix2pix~\cite{pix2pix} &0.031 &10.784 & 27.60 & 0.864 &54.41 &6.1\\
        SPA-GAN~\cite{SPA-gan-2020} &0.045 & 18.085 & 24.78 & 0.754 &0.21 &15.2\\
		DSen2-CR~\cite{meraner2020cloud} & 0.031  & 9.472  &27.76 &0.874 &18.92 &1240.2\\
		GLF-CR~\cite{xu2022glf} & 0.028  & 8.981  &28.64 &0.885&14.82&250.0\\
		UnCRtainTs L2~\cite{ebel2023uncrtaints} & 0.027	&8.320	&28.90 	&0.880 & 0.56 & 37.1\\
		ACA-CRNet~\cite{huang2024attentive} & 0.025	& 7.770	& 29.78	& 0.896 &20.39 &1422.0 \\
        \midrule
            Ours & \textbf{0.024} & \textbf{7.484}	& \textbf{29.97} & \textbf{0.902} & 6.93 & 53.3\\

		\bottomrule
		\end{tabular}}

	\label{tab:sen12ms-results}%
\end{table}%

We also conduct experiments on SEN12MS-CR~\cite{sen12mscr}, a large-scale multi-spectral remote sensing dataset with 122,218 training samples, to evaluate our model's performance. Table~\ref{tab:sen12ms-results} reports the experimental results, including attention-based methods such as SPA-GAN~\cite{SPA-gan-2020}, CLF-CR~\cite{xu2022glf}, UnCRtainTs~\cite{ebel2023uncrtaints}, and ACA-CRNet~\cite{huang2024attentive}. Our method achieves a PSNR of 29.97db, an SSIM of 0.902, and a SAM of 7.484. Compared to the existing SOTA method, ACA-CRNet, our ATT-CR reduces the SAM by 0.28 and improves PSNR by 0.19dB while using only 34\% of the parameters and 4\% of the computational cost. The superiority of our model stems from its ability to capture pixel-level long-range dependency and FSGM's adaptive feature selection, which significantly enhances feature representation capability. In contrast, ACA-CRNet's attention module, with $\mathcal{O}(N^2)$ computational complexity, is restricted to the deeper layers to minimize computational costs, which leads to weaker feature representation capabilities.

The last two columns of Table~\ref{tab:main-results} and Table~\ref{tab:sen12ms-results} provide a comparison of the parameter counts and FLOPs across the evaluated models, where these metrics are measured at inputs of $256\times 256$. Although pix2pix and SPA-GAN have relatively fewer FLOPs, their performance is notably inferior. In contrast, our model achieves superior performance while significantly reducing computational costs compared to SOTA methods. For example, on the T-CLOUD, RICE1, and RICE2 datasets, our model reduces the parameter count by 73.4\% and the FLOPs by 65.5\% compared to CR-former. On the SEN12MS-CR dataset, our model requires only 34\% of the parameters and 4\% of the computational cost compared to ACA-CRNet. These quantitative results demonstrate that our approach effectively balances performance and computational complexity.

\subsubsection{\textbf{Qualitative Comparisons}}
\begin{figure*}[!t]
	\centering
	\includegraphics [width=6.4in]{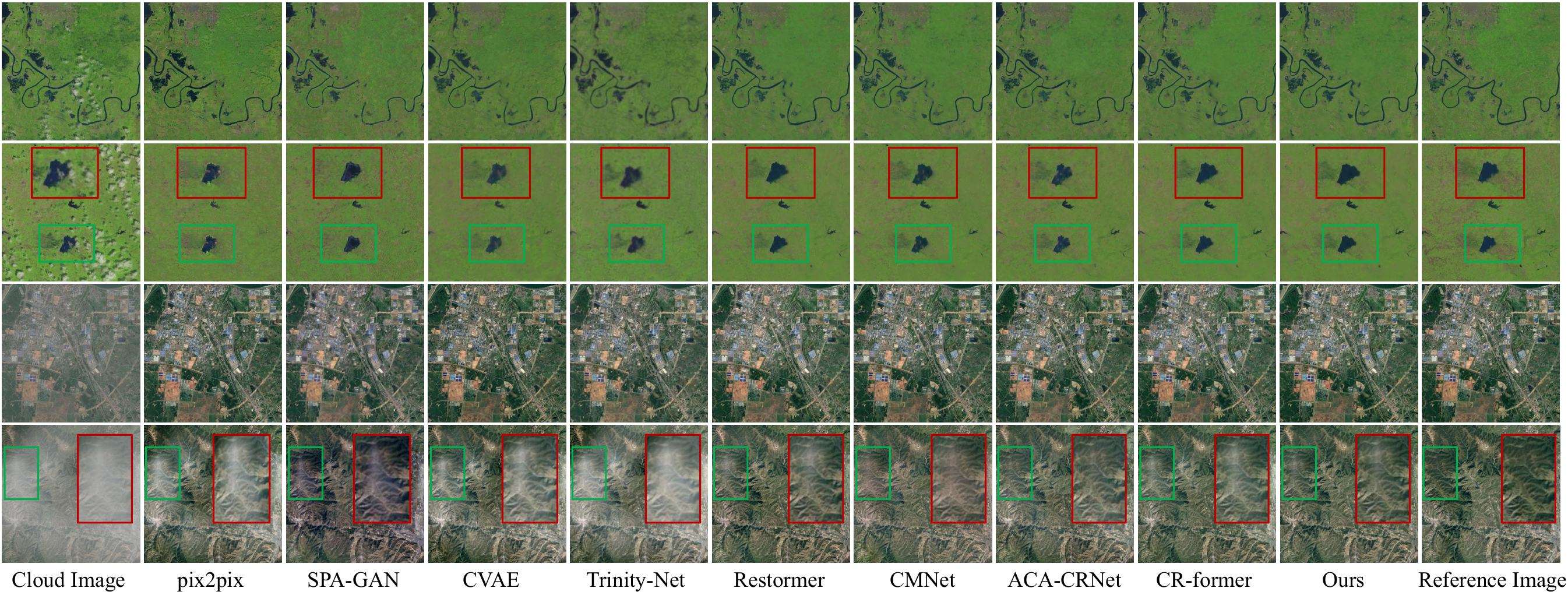}
\caption{Illustration of cloud removal results on the RICE dataset. The first two rows correspond to RICE2, while the last two belong to RICE1. Cloudy input images are shown in the first column, and the reference (ground truth) images are provided in the last column. The intermediate columns display the outputs from several baseline models and our proposed method. The areas marked by the green box are enlarged within the red box for better clarity. }
	\label{fig:rice-vis}
\end{figure*}

\begin{figure*}[!t]
	\centering
	\includegraphics [width=6.4in]{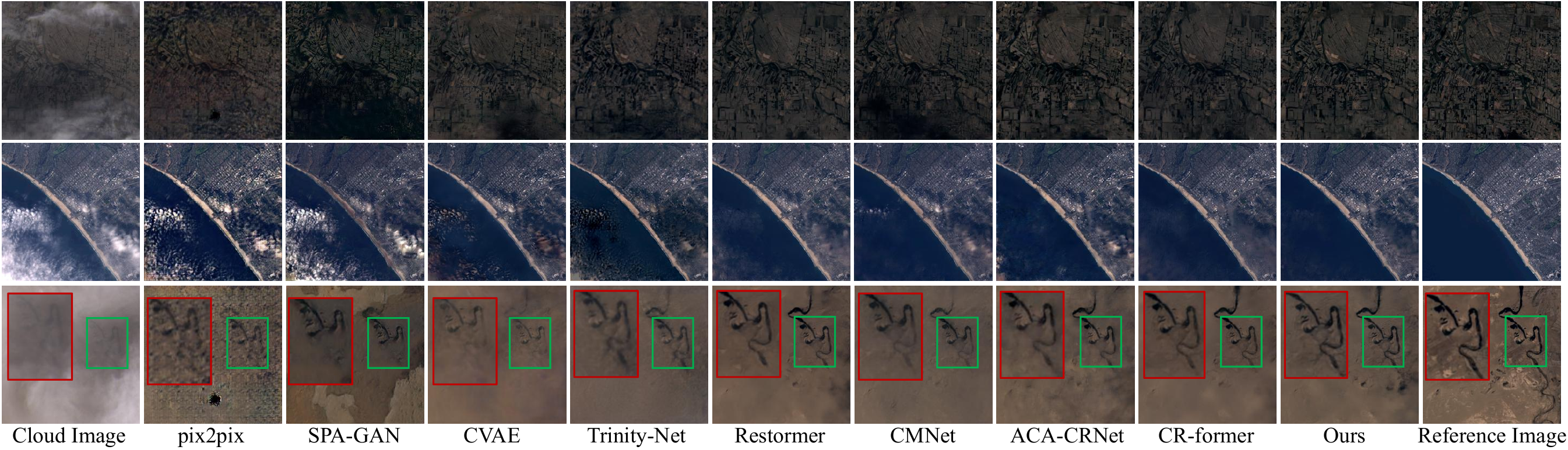}
\caption{Illustration of cloud removal results on the T-CLOUD dataset. Cloudy input images are shown in the first column, and the reference (ground truth) images are provided in the last column. The intermediate columns display the outputs from several baseline models and our proposed method. The areas marked by the green box are enlarged within the red box for better clarity.}
	\label{fig:tcloud-vis}
\end{figure*}

\begin{figure*}[!t]
	\centering
	\includegraphics [width=6.4in]{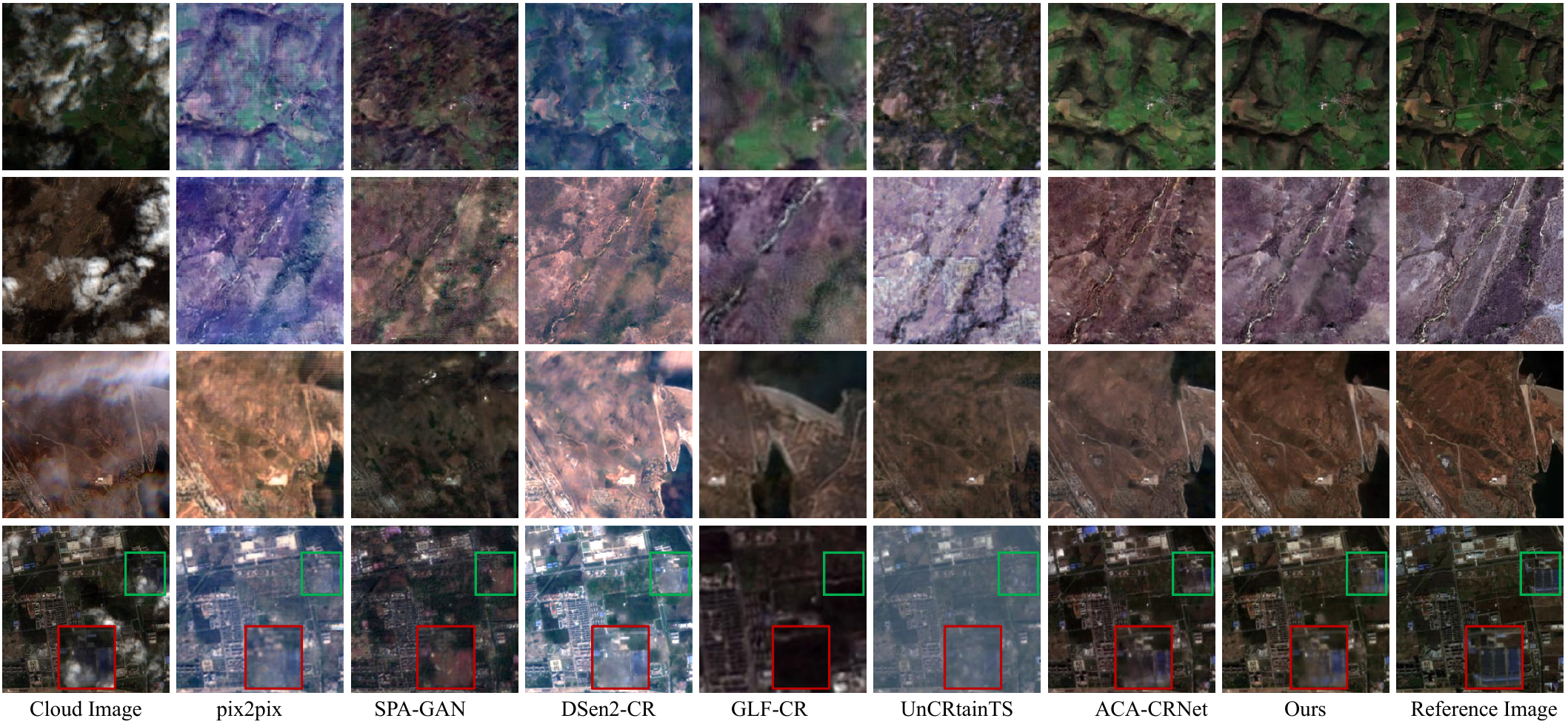}
\caption{Illustration of cloud removal results in the RGB channels for the SEN12MS-CR dataset. Cloudy input images are shown in the first column, and the reference (ground truth) images are provided in the last column. The intermediate columns display the outputs from several baseline models and our proposed method. The areas marked by the green box are enlarged within the red box for better clarity.}
	\label{fig:sen12ms-vis}
\end{figure*}

In Figs.~\ref{fig:rice-vis}, \ref{fig:tcloud-vis}, and \ref{fig:sen12ms-vis}, we show some qualitative results generated by the baseline models on the RICE1, RICE2, T-CLOUD, and SEN12MS-CR. In the diverse cloud coverage scenarios, compared to other methods, our model performs well with superior clarity and high-quality detail restoration.

Fig.\ref{fig:rice-vis} presents qualitative results on the RICE1 and RICE2 datasets. The first two rows display restored images from the thick cloud dataset RICE2, while the last two rows show images from the thin cloud dataset RICE1. Among the evaluated methods, the CNN-based pix2pix\cite{pix2pix} suffers from grid artifacts, and SPA-GAN~\cite{SPA-gan-2020} introduces noticeable color distortions. Models such as CVAE~\cite{cvae-DingZX22}, Trinity-Net~\cite{chi2023trinity}, and CMNet~\cite{cmnet} struggle with large cloud-covered regions, leaving residual clouds and failing to recover crucial features like rivers and mountain ridges, as seen in the second and last rows of Fig.\ref{fig:rice-vis}. This limitation arises from their reliance on Swin\cite{swin-transformer} or ViT~\cite{VIT}, which struggle with restricted receptive fields or coarse feature representations for global context modeling. In contrast, our model effectively removes clouds and preserves the perceptual integrity of the image by leveraging pixel-level long-range dependencies. The visualization results confirm that our method performs effectively on both thin and thick cloud scenarios.

To further assess our model's performance in real-world cloud scenarios, we conducted experiments on the T-CLOUD dataset, which includes real-world cloudy and cloud-free images captured from the same locations at different times. Fig.\ref{fig:tcloud-vis} presents visualization results on the T-CLOUD test dataset. Previous state-of-the-art methods, such as Restormer\cite{restormer} and CR-Former, successfully remove clouds but exhibit noticeable blurring along the restored edges, as shown in the last row of Fig.~\ref{fig:tcloud-vis}. This issue arises from neglecting the influence of cloudy pixels in attention computation and the low-rank limitation in linear attention, leading to visual artifacts. In contrast, our model achieves superior performance, effectively removing clouds while preserving fine-grained textures and high fidelity. The restored image clearly demonstrates the effectiveness and robustness of our method in real-world cloud conditions.

Furthermore, to validate ATT-CR’s robustness, we conduct experiments on the large-scale multispectral SEN12MS-CR dataset and present qualitative comparison results in the RGB channels. As illustrated in Fig.~\ref{fig:sen12ms-vis}, compared to competing models, which often introduce noticeable artifacts or fail to remove cloud coverage, our approach restores images with minimal distortion and accurate color rendition. For instance, the previous state-of-the-art model, ACA-CRNet\cite{huang2024attentive}, struggles with large cloud-covered areas and fails to recover critical elements. In contrast, our model achieves superior image restoration with higher clarity and fine-detail preservation. These results highlight the outstanding performance and robustness of ATT-CR on large-scale datasets.


These qualitative comparisons demonstrate the effectiveness and robustness of ATT-CR in addressing challenges posed by various cloudy scenes. It produces high-fidelity, cloud-free images by effectively modeling pixel-level long-range dependencies, without restricting receptive fields, and adaptively selecting features to reduce disturbance from cloudy pixels.

\subsection{Ablation Studies}
This section presents a comprehensive analysis of each module designed in ATT-CR to illustrate the effectiveness of our approach. The model is trained on $256  \times 256$ pixel patches cropped from the input cloudy images, with all evaluations conducted on the RICE1 or RICE2 datasets. 

\begin{table}[!tbp]
	\centering
	\caption{Effectiveness of key designs TAN, FSGM, and MS-Tokens in ATT-CR. $\downarrow$: Lower values indicate better performance, $\uparrow$: Higher values indicate better performance.}
    \renewcommand\arraystretch{1}
    \setlength{\tabcolsep}{0.5mm}{
	\begin{tabular}{c|ccc|ccccc}
    \toprule
    \multirow{1}[1]{*}{\makecell[c]{\textbf{Dataset}}} & \multirow{1}[1]{*}{\makecell[c]{\textbf{TAN}}} & \multirow{1}[1]{*}{\makecell[c]{\textbf{MS-Tokens}}} &\multirow{1}[1]{*}{\makecell[c]{\textbf{FSGM}}} &  \multirow{1}[1]{*}{\makecell[c]{\textbf{MAE} $\downarrow$}} &  \multirow{1}[1]{*}{\makecell[c]{\textbf{SAM} $\downarrow$ } }&  \multirow{1}[1]{*}{\makecell[c]{\textbf{PSNR}$\uparrow$}} &  \multirow{1}[1]{*}{\makecell[c]{\textbf{SSIM}$\uparrow$} } \\
    \midrule
     \multirow{4}[1]{*}{\makecell[c]{\textbf{RICE1}}} & \checkmark & \checkmark & \checkmark & 0.0135	& 0.9141	& 36.83	 &0.9629 \\

     & \ding{55} & \checkmark  & \checkmark & 0.0142 & 0.9183	&36.60	&0.9627 \\
      &  \checkmark  &\ding{55} &\checkmark   & 0.0139 &0.9197 & 36.69	&0.9626 \\

       & \checkmark  & \checkmark &\ding{55} &0.0141	& 0.9456 &36.63	&0.9621 \\
       & \ding{55}  & \ding{55} &\ding{55} & 0.0148 & 0.9452 & 36.20 &0.9617\\
       
       \midrule

        \multirow{4}[1]{*}{\makecell[c]{\textbf{RICE2}}} & \checkmark & \checkmark & \checkmark & 0.0149	& 1.166	& 36.72	 &0.9183 \\
   
     & \ding{55} & \checkmark  & \checkmark & 0.0155 & 1.206	&36.44	&0.9176 \\
      &  \checkmark  &\ding{55} &\checkmark  &0.0153	& 1.165 &36.64	&0.9179 \\ 
    
       & \checkmark  & \checkmark &\ding{55} & 0.0152	&1.217	&36.44	&0.9168 \\
       & \ding{55}  & \ding{55} &\ding{55} &0.0161	& 1.249 &36.11	&0.9164 \\
    \bottomrule
    \end{tabular}
    }
  \label{tab:ablation}%
\end{table}

Our model consists of the primary designs of TAN, FSGM, and Multi-scale Tokens (MS-Tokens), which achieve a balance between high-quality feature representation and computational efficiency. TAN effectively models the long-range dependency with $\mathcal{O}(N)$ computational complexity and mitigates the low-rank bottleneck inherent in linear attention, while MS-Tokens capture local context at multiple scales to enhance feature diversity. FSGM adaptively selects features to mitigate the impact of cloudy regions, further enhancing the model's robustness. To evaluate the effectiveness of these designs, we conduct experiments by sequentially removing each component from the network.

\begin{itemize}
\item {\textbf{TAN}: As shown in Table~\ref{tab:ablation}, removing TAN and replacing it with the original linear attention without the triangular matrix segmentation (w/o TAN) results in a PSNR drop of 0.23dB and 0.28dB on the RICE1 and RICE2 datasets, respectively. Fig.~\ref{fig:ablation-vis} provides further visual evidence of this ablation experiment. Comparing columns (a) and (e) in Fig.~\ref{fig:ablation-vis}, we observe that the absence of TAN leads to incomplete restoration of the cloud-affected region and certain ground details to be lost. This degradation is attributed to the low-rank nature of the original linear attention, which leads to homogenized attention weights, making the aggregated features indistinguishable. Consequently, the restored image in (a) appears ambiguous and lacks detail compared to (e). These results highlight TAN’s effectiveness in mitigating low-rank limitations of linear attention, leading to more accurate restorations.}

\begin{figure*}[!t]
	\centering
\includegraphics [width=5.2in] {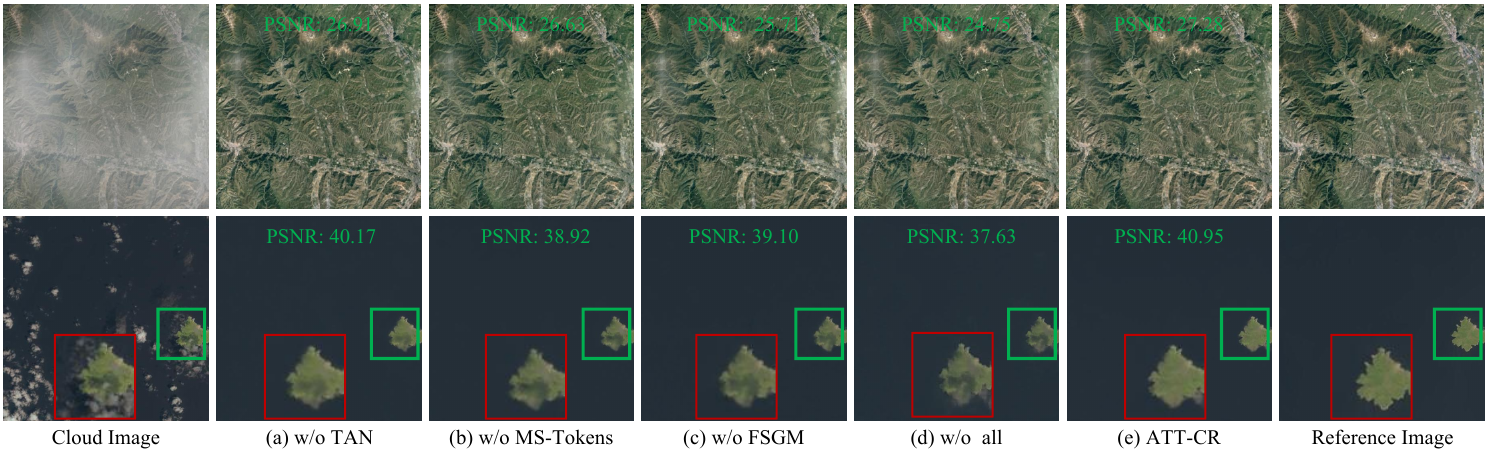}
\caption{Ablation visualization results from the RICE1 and RICE2 datasets. (a) removing the TAN; (b) removing the MS-Tokens; (c) removing the FSGM; (d) removing all designs of TAN, MS-Tokens, and FSGM; (e) our full model: ATT-CR.}
	\label{fig:ablation-vis}
\end{figure*}

\item{\textbf{MS-Tokens}: MS-Tokens enables the model to capture ground objects at multiple scales, which is crucial for preserving texture details and boundaries. Table~\ref{tab:ablation} shows that removing MS-Tokens results in PSNR decrease of 0.14dB and 0.08dB on RICE1 and RICE2, respectively. Additionally, the comparison between (b) and (e) in Fig.~\ref{fig:ablation-vis} illustrates that MS-Tokens help preserve clear edges and fine-grained information of the restored image. These results demonstrate the effectiveness of MS-Tokens in improving image quality.}

 \item{\textbf{FGSM}: As seen in Table~\ref{tab:ablation}, removing FSGM results in PSNR reductions of 0.2dB on RICE1 and 0.28dB on RICE2, indicating a significant performance drop. Additionally, in Fig.~\ref{fig:ablation-vis} (comparison between (c) and (e)), the recovered image (c) exhibits noticeable color inconsistencies and blurriness, particularly in the cloud-covered regions, compared to the full model in (e). These qualitative and quantitative results confirm that FSGM plays a crucial role in enhancing image clarity.}
\end{itemize}

When all designed components are removed, the model's performance significantly degrades, with PSNR drops of 0.63dB and 0.61dB on RICE1 and RICE2, respectively. As shown in Fig.~\ref{fig:ablation-vis}, comparing (d) and (e), the restored image exhibits noticeable blurriness and missing details. These experimental results highlight the individual contributions of each component and the synergistic effect of combining them.

\begin{table}[tb]
	\centering
	\caption{ Comparison with other efficient attention methods on dataset RICE2. Replace the TAN with other efficient attention. $\downarrow$: Lower values indicate better performance, $\uparrow$: Higher values indicate better performance.}
    \renewcommand\arraystretch{1.1}
    \setlength{\tabcolsep}{1.1mm}{
	\begin{tabular}{c|cc|cc}
    \toprule
     \multirow{3}[1]{*}{\makecell[c]{\textbf{Method}}}  & \multirow{3}[1]{*}{\makecell[c]{\textbf{PSNR}$\uparrow$}} &  \multirow{3}[1]{*}{\makecell[c]{\textbf{SSIM}$\uparrow$} } & \multicolumn{2}{c}{\textbf{Overhead}} \\
     &&&Param & FLOPs \\
     &&& (M) & (G) \\
    \midrule
		Ours (TAN)   & \textbf{36.72}	 &\textbf{0.9183}  &6.91 &53.03 \\
 
    	TAN$\rightarrow$Swin~\cite{swin-transformer}  & 36.17 & 0.9179 & 6.97  & 58.65\\
            TAN$\rightarrow$Linear SRA~\cite{wang2022pvt}  & 36.60 & 0.9177 & 7.60 & 46.27\\
		TAN$\rightarrow$MDTA~\cite{restormer}  & 36.45  & 0.9177 &  6.91 & 56.31\\
		TAN$\rightarrow$EA~\cite{EA-ShenZZY021}  &36.37   & 0.9177 & 6.91 & 56.31 \\
		TAN$\rightarrow$ ELU LA~\cite{linear_transformer} & 36.44 &0.9176  & 6.91 &56.35\\
        TAN $\rightarrow$FT-Attention~\cite{wu2024cr}  & 36.50  & 0.9180 &6.91 &56.35 \\

    \bottomrule
    \end{tabular}
    }
  \label{tab:ablation_attn}%
\end{table}
\subsection{Comparison with Other Efficient Attention}

To ensure fairness in comparisons with other efficient attention mechanisms, we replace the TAN attention module in our network with other representative efficient attention modules while keeping the MS-Tokens and FSGM in the network architecture. Table~\ref{tab:ablation_attn} presents the experimental results on the RICE2 dataset, including parameter counts and FLOPs, which provides a comprehensive comparison of performance and computational costs. Specifically, we compared our TAN with five previous efficient attention designs, including Swin attention~\cite{swin-transformer}, Linear Spatial Reduction Attention (Linear SRA)~\cite{wang2022pvt}, Multi-Dconv Transposed Attention (MDTA)~\cite{restormer}, Efficient Attention (EA)~\cite{EA-ShenZZY021}, ELU Linear Attention (ELU LA)~\cite{linear_transformer}, and FT-Attention~\cite{wu2024cr}.

Among these models, Swin~\cite{swin-transformer} employs shifted window attention to model long-range dependencies but with regional receptive fields and complex shifting operations. Linear SRA~\cite{wang2022pvt} reduces attention tokens by applying average pooling on keys and values, but it sacrifices some fine-grained pixel-level information. MDTA~\cite{restormer} captures pixel-level long-range dependencies via channel attention, yet loses spatial information. EA~\cite{EA-ShenZZY021}, ELU LA~\cite{EA-ShenZZY021}, and FT-Attention~\cite{wu2024cr} utilize the carefully designed mapping function to approximate the softmax attention with $\mathcal{O}(N)$ computational complexity, but exacerbate the low-rank limitation inherent in multi-head attention, restricting feature diversity. In contrast, our TAN not only models pixel-level long-range dependencies with $\mathcal{O}(N)$ computational complexity, without restricting the receptive fields, but also mitigates the low-rank limitation. Table~\ref{tab:ablation_attn} shows that our TAN significantly outperforms all other efficient designs while maintaining similar parameter counts and FLOPs, highlighting that our method achieves a superior balance between high expressive capability and computational efficiency.

\subsection{Visualization of Gated Module Values}
\label{Visualization of Gated values}
The Feature Selection Gating Module (FSGM) adaptively selects features for each channel at every spatial location, mitigating the disturbance caused by cloudy features in attention computation. This facilitates the propagation of valid information to subsequent layers and enhances the model’s robustness against cloud contamination. To validate the effectiveness of FSGM and provide an interpretable analysis, we visualize the learned gating values within the cloud removal network.
\begin{figure*}[!t]
	\centering
	\includegraphics [width=4.6in]{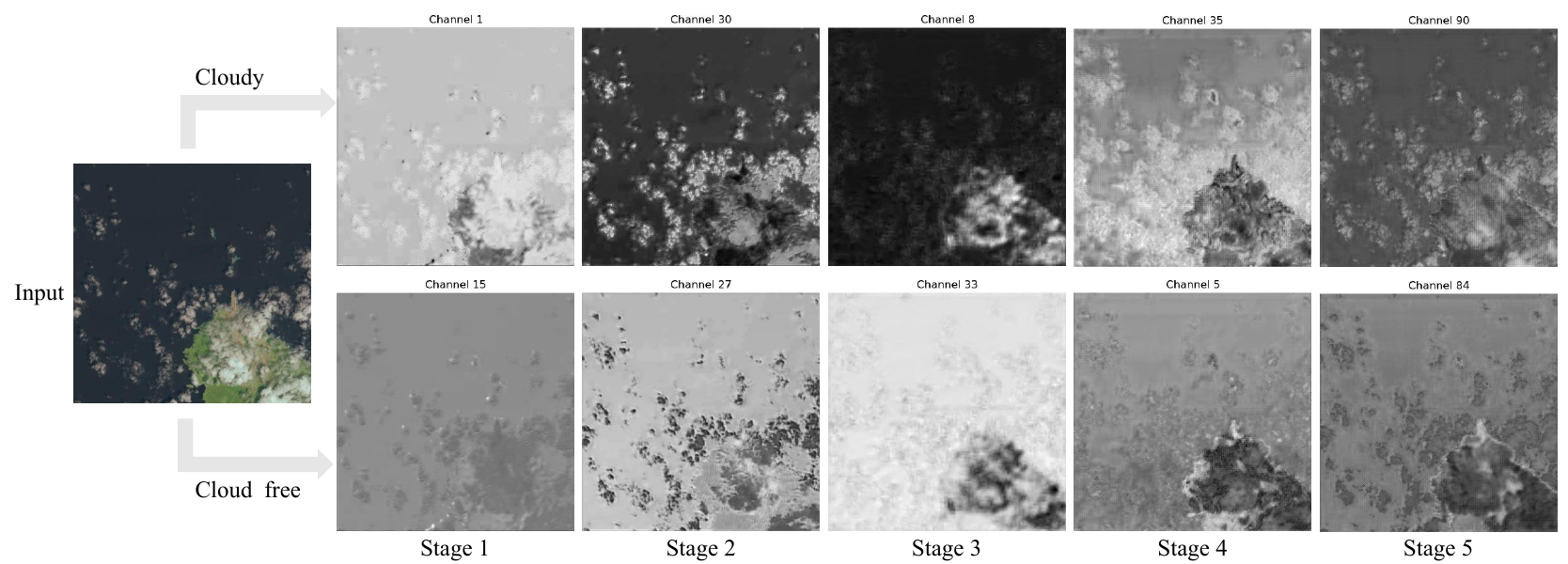}
    \caption{
    The visualization of the output values from the learned FSGM, based on results from the RICE2 dataset, shows varying characteristics across different channels and locations at each stage. These gating values effectively differentiate between cloudy and cloud-free features, facilitating the propagation of valid features to subsequent layers.
    }
\label{fig:gate-channel-vis}
\end{figure*}

The unnormalized gating values produced by the FSGM module in the final block of each stage are visualized in Fig.~\ref{fig:gate-channel-vis}. These values reveal distinct patterns, with higher gating values indicating the focus of each channel. Specifically, some channels concentrate on cloudy regions to capture missing information, as certain bands can penetrate the clouds and retain specific ground details. Other channels focus on cloud-free areas to filter out contaminated features and preserve informative, clean features. This visualization demonstrates that the model adaptively selects features, distinguishing valuable information from cloud-contaminated regions.

\begin{figure}[!ht]
	\centering
	\includegraphics[width=3.3in]{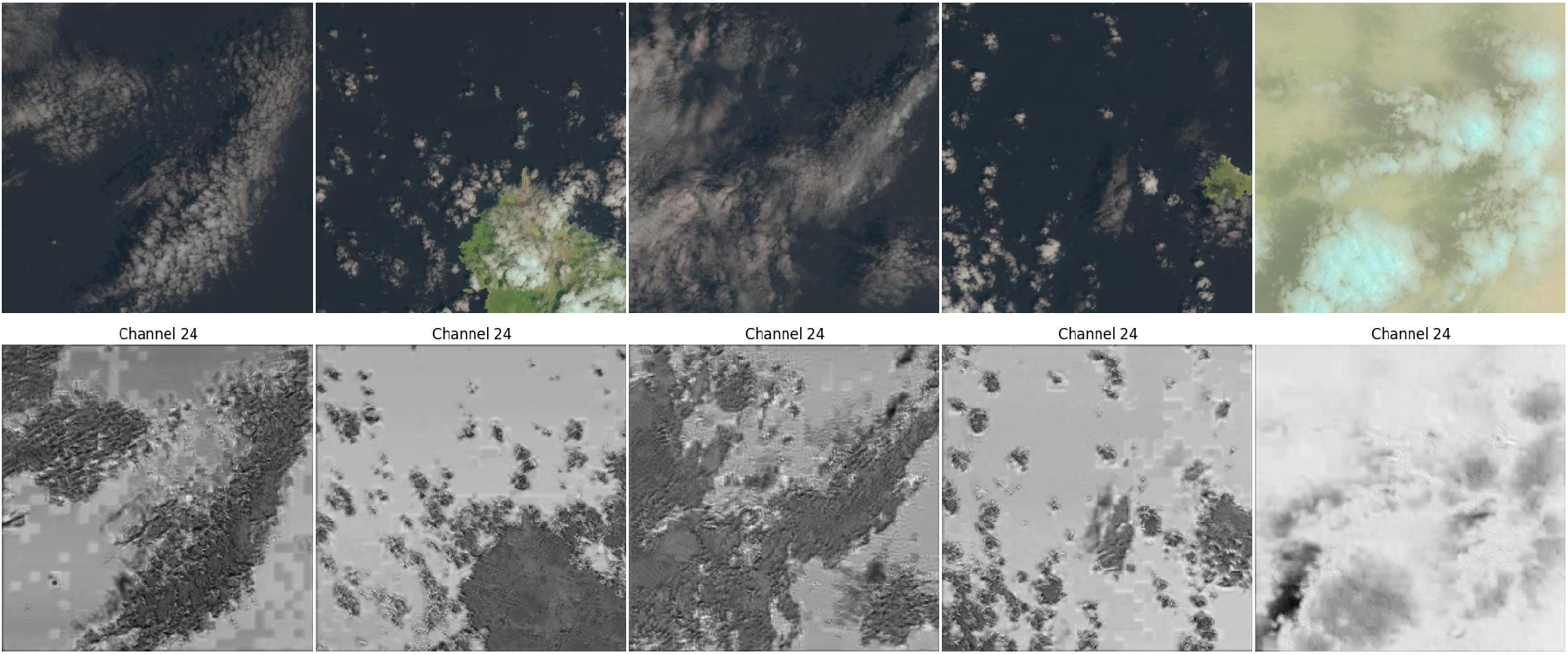}
    \caption{Selected gating channel outputs from FSGM at stage 2. The first row shows the input cloudy images, and the second row displays the gating channel outputs from FSGM. Darker regions indicate suppressed (cloudy) areas; brighter regions indicate preserved or enhanced features. These results show the adaptive spatial response of FSGM to different cloud patterns.} 
	\label{fig:spatial-choose}
\end{figure}

\begin{figure*}[!ht]
	\centering
	\includegraphics[width=6.4in]{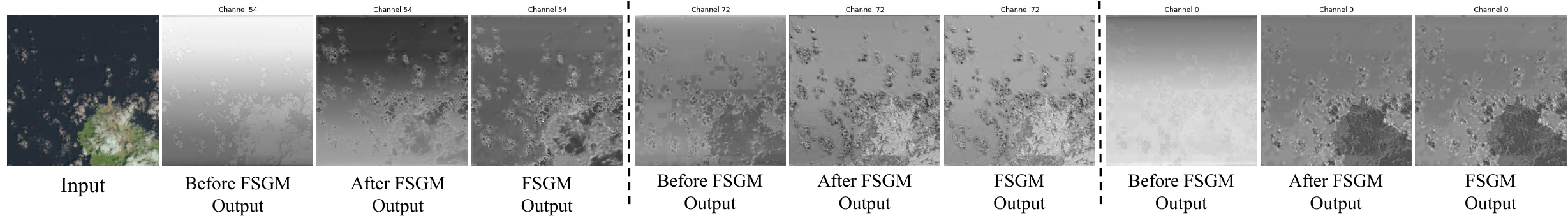}
    \caption{Visualization of selected channel outputs before and after FSGM and FSGM gating values. Darker regions indicate suppressed areas; brighter regions indicate preserved or enhanced features.} 
	\label{fig:channel-choose}
\end{figure*}

\revise{Additionally, we visualize the output of the selected channel from FSGM at Stage 2 for various cloudy inputs in Fig.~\ref{fig:spatial-choose}. The results show that the spatial distribution of gating responses adapts to varying cloud coverage. Specifically, suppressed activation is observed in cloud-covered regions, while enhanced responses appear in clean areas, illustrating how FSGM’s spatial selection dynamically responds to degradation in the input images.
Fig.~\ref{fig:channel-choose} further compares feature maps before and after FSGM. Without FSGM, cloudy regions retain high activation values, but after applying FSGM, these regions are significantly suppressed. The gated output highlights how FSGM selectively enhances valid features while suppressing cloud-related noise, providing strong evidence of its role in cloud suppression and feature refinement.}

\begin{figure*}[!th]
	\centering
	\includegraphics [width=5.2in]{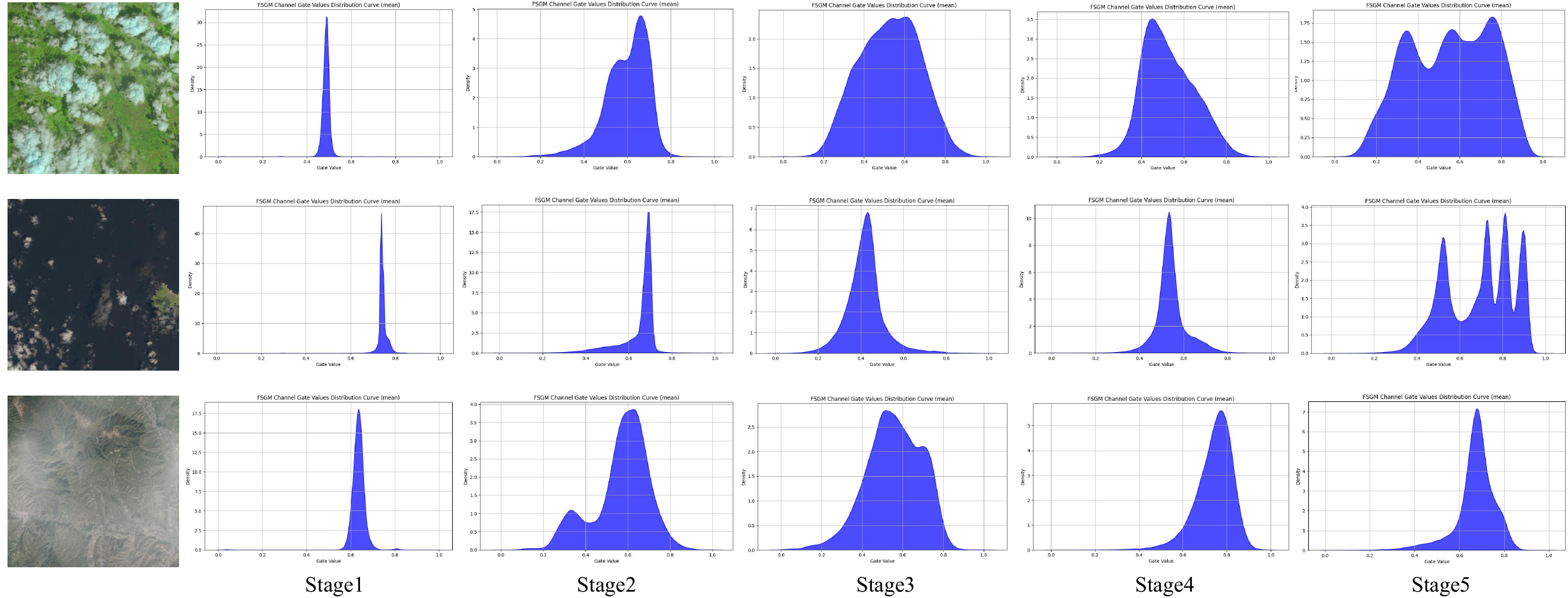}
    \caption{
   FGSM gate value distributions (channel-wise mean) across five network stages for large-scale thick cloud (top), small-scale thick cloud (middle), and thin cloud (bottom). Gate values are low in early stages for dense clouds due to severe occlusion, and higher for thin and small-scale thick clouds owing to the availability of clean regions. As the network deepens, the distributions gradually broaden as informative features are recovered. This demonstrates FGSM’s adaptive modulation based on cloud density and coverage.
    }
\label{fig:gate-distribution_vis}
\end{figure*}

To further investigate how the Feature Selection Gating Module (FSGM) adapts to varying cloud conditions across different stages, we visualize the mean gate values (averaged across all channels) at the final block of each stage for representative thin and thick cloud images, as shown in Fig.~\ref{fig:gate-distribution_vis}. The results reveal clear trends: for thin clouds, the gating activations are consistently high and concentrated across all stages, indicating confident feature selection under light occlusion. In contrast, thick cloud images exhibit more complex patterns. For large-scale occlusions, early-stage gate values are low and focused, while small-scale occlusions yield higher early-stage activations due to the availability of clean regions. As the network deepens (Stage 2–4), gate distributions broaden, reflecting the progressive recovery of meaningful features. At the final stage, gate values display a multi-modal distribution, which can be attributed to the model's residual learning structure designed to capture the difference between the degraded input and the clean target. As a result, the final stage primarily focuses on the scattered degraded regions. These observations validate that FSGM dynamically adjusts feature selection in response to cloud density and spatial distribution.

\subsection{Model Complexity Analysis}

\begin{table}[!t]
	\centering
	\caption{Comparison of FLOPs and Inference Times Across Different Models on GPU and CPU for Various Input Resolutions.}
    \resizebox{1\columnwidth}{!}{
    \renewcommand\arraystretch{0.9}
    \setlength{\tabcolsep}{0.65mm}
	\begin{tabular}{ c c c c c}
    \toprule
    \multirow{1}[1]{*}{\textbf{Input Sizes}}& \multirow{1}[1]{*}{\textbf{Model}} & 
   \multirow{1}[1]{*}{ \textbf{FLOPs} (G) $\downarrow$} & \multirow{1}[1]{*}{\textbf{GPU} (ms) $\downarrow$} & \multirow{1}[1]{*}{\textbf{CPU} (ms) $\downarrow$} \\ [4pt]
    \midrule
    \multirow{5}[1]{*}{\makecell[c]{ $32 \times 32$}} 
    & {CVAE~\cite{cvae-DingZX22}} & 0.59 & 4.31 & 14.78 \\
     & {Restormer~\cite{restormer}} & 2.42 & 39.47 & 160.42  \\
     & {CMNet~\cite{cmnet}} & 3.68 & 90.16 & 279.94 \\
    & {ACA-CRNet~\cite{huang2024attentive}} & 17.98 & 5.61 & 76.20 \\
     & {CR-former~\cite{wu2024cr}} & 2.42 & 60.63 & 206.78  \\
    &  Ours &  0.83 &  32.21 &  111.08 \\
   \midrule
    \multirow{5}[1]{*}{\makecell[c]{$64 \times 64$ }} 
      & CVAE~\cite{cvae-DingZX22} & 2.32 & 4.45 & 22.50  \\
      & Restormer~\cite{restormer}& 9.68 & 40.32 & 254.75\\
      &  CMNet~\cite{cmnet}& 14.73 & 97.47 & 420.73   \\
      & ACA-CRNet~\cite{huang2024attentive}& 72.74 & 10.45 & 256.67 \\
      & CR-former~\cite{wu2024cr} & 9.68 & 64.79 & 388.59  \\
      & {Ours} & 3.33 & 33.82 &215.85\\
   
    \midrule
       \multirow{5}[1]{*}{\makecell[c]{$128 \times 128$ }} 
        &CVAE~\cite{cvae-DingZX22} & 9.24 & 5.02 & 61.71  \\
        & Restormer~\cite{restormer}& 38.72 & 42.64 & 610.53  \\
        & CMNet~\cite{cmnet}& 58.93 & 103.72 & 795.38  \\
        & ACA-CRNet~\cite{huang2024attentive} & 311.30 & 40.26 & 1558.25\\
        & CR-former~\cite{wu2024cr} & 38.74 & 64.96 & 797.24  \\
        & Ours & 13.34 &40.75 & 542.62 \\
 \midrule
     \multirow{5}[1]{*}{\makecell[c]{$256 \times 256$ }} 
    &CVAE~\cite{cvae-DingZX22} & 15.42 & 15.10 & 209.12  \\
    & Restormer~\cite{restormer} & 154.88 & 84.74 & 2068.84  \\
     & CMNet~\cite{cmnet}& 236.00 & 171.99 & 2318.31  \\
     & ACA-CRNet~\cite{huang2024attentive} & 1456.13 & 223.74 & 17912.23\\
   & CR-former~\cite{wu2024cr} & 154.97 &109.01 & 1938.65 \\
    & Ours & 53.34  &108.69 & 1745.74 \\
    \bottomrule
    \end{tabular}
    }
  \label{tab:complexity}%
\end{table}
In this section, we provide a detailed analysis of our model’s computational costs. Table~\ref{tab:complexity} reports the FLOPs and inference times of our method and several representative Transformer-based baselines across input sizes ranging from $32 \times 32$ to $256 \times 256$. \revise{All tests were conducted on a single NVIDIA RTX 3090 GPU and dual Intel Xeon Silver 4214R CPUs (48 logical cores in total)}.

Our model demonstrates favorable computational efficiency across different input sizes. At higher resolution input ($256 \times 256$), our FLOPs (53.34G) are significantly lower than those of CMNet (236.00G), Restormer (154.60G), and ACA-CRNet (1456.13G), while being only slightly higher than CVAE (15.42G), which, however, yields inferior restoration performance (PSNR scores 3.13 dB and 3.10 dB lower than ours on RICE1 and RICE2, respectively).

In terms of inference speed, our model achieves a GPU runtime of 108.69 ms at $256 \times 256$, which, while not the fastest due to the serial nature of triangular attention, remains competitive and notably faster than ACA-CRNet (223.74 ms) and CMNet (171.99 ms). On CPU, our method demonstrates even more significant advantages, with an inference time of 1745.74 ms at $256 \times 256$, outperforming CR-former (1938.65 ms) by approximately 10\%, CMNet (2068.84 ms) by 15\%, and ACA-CRNet (2311.92 ms) by over 90\%.

These results demonstrate that our method achieves an excellent trade-off between efficiency and performance. Its consistently low FLOPs and runtime, particularly on CPU, make it a strong candidate for practical deployment scenarios with limited computational resources.

\subsection{Ablation on Hyperparameters}

 For selecting multi-scale convolution kernel sizes, we conducted ablation experiments with the following configurations: Single-scale: kernel size = [1], [3], [5]; Dual-scale: kernel sizes = [1, 3], [3, 5]; and Triple-scale: kernel sizes = [1, 3, 5].
 
 As shown in Table~\ref{tab:kernel}, the multi-scale configurations consistently outperform their single-scale counterparts in both PSNR and SSIM metrics, with the triple-scale variant achieving the highest scores. This demonstrates that incorporating multi-scale receptive fields can effectively enhance restoration quality. To balance model efficiency and inference speed, we adopt a hybrid strategy during training: using the [3, 5] kernel configuration for higher-resolution stages (e.g., stages 1 and 5), and the [1, 3] configuration for lower-resolution stages (e.g., stage 3).

\begin{table}[tbp]
	\centering
	\caption{Ablation Results on Multi-Scale Kernel Configurations on the RICE2 dataset.}
    \renewcommand\arraystretch{1}
    \setlength{\tabcolsep}{3.6mm}{
	\begin{tabular}{c|ccccc}
    \toprule
    \multirow{1}[1]{*}{\makecell[c]{\textbf{Kernel Size}}} &\multirow{1}[1]{*}{\makecell[c]{\textbf{MAE} $\downarrow$}} &  \multirow{1}[1]{*}{\makecell[c]{\textbf{PSNR}$\uparrow$}} &  \multirow{1}[1]{*}{\makecell[c]{\textbf{SSIM}$\uparrow$} } \\
    \midrule
     \multirow{1}[1]{*}{\makecell[c]{\textbf{1}}} &0.0153  &36.64 &0.9179\\
      \textbf{3}  & 0.0151 & 36.64  & 0.9180\\
      \textbf{5}  &0.0153& 36.52 & 0.9178\\
     \textbf{(1,3)} &0.0150 &  36.65 & 0.9179\\
     \textbf{(3,5)}  & 0.0150  & 36.66 & 0.9182\\
      \textbf{(1,3,5)} & 0.0147 & 36.77 & 0.9182 \\
    \bottomrule
    \end{tabular}
    }
  \label{tab:kernel}%
\end{table}

\section{Discussion} \label{Discussion}
While the proposed ATT-CR framework achieves strong performance in cloud removal and exhibits competitive inference efficiency, several insights and limitations emerged from our analysis that deserve further discussion.

The Feature Selection Gating Module (FSGM) is central to the model’s adaptive capability. Gate value visualizations show that FSGM dynamically adjusts feature selection across stages and cloud conditions. However, due to varying channel dimensions and the black-box nature of deep networks, interpreting fine-grained channel behaviors remains challenging. Enhancing the interpretability of this gating process is a promising direction for future research. Additionally, due to the serial nature of triangular attention, we plan to explore custom operators to accelerate triangular attention and further enhance runtime efficiency.

\section{Conclusion} \label{conclusion}
This paper proposes the Adaptive Triangular Transformer for Cloud Removal (ATT-CR), which is designed to efficiently and effectively restore cloudy remote sensing images. Our model comprises two key modules: the Triangular Attention (TAN) and the Feature Selected Gating Module (FSGM). TAN captures the pixel-level long-range dependency with $\mathcal{O}(N)$ computational complexity and alleviates the low-rank limitation inherent in linear attention. FSGM adaptively selects important features across spatial and channel dimensions, enhancing robustness to cloudy contamination. The experimental results indicate that ATT-CR improves both accuracy and computational efficiency, demonstrating its potential as a promising method for cloud removal in remote sensing applications.

\revise{Furthermore, while ATT-CR is primarily designed for cloud removal, its core components—Triangular Attention (TAN) and Feature Selection Gating Module (FSGM)—are inherently generalizable to a broader range of remote sensing restoration tasks. In particular, the ability of TAN to capture fine-grained long-range dependencies and the adaptive modulation provided by FSGM make the model well-suited for handling other types of occlusion and degradation, such as haze, shadow, and atmospheric interference. These forms of degradation also involve large-scale structural occlusion and ambiguous texture, for which our architecture's design can be readily adapted. In addition, incorporating semantic priors from large vision foundation models, which have shown strong transferable representations in related vision tasks~\cite{liu2024semantic, liu2025mind, qi2026patchcue, yang2026shaping, liu2026structured}, offers another promising avenue for improving robustness under complex and diverse degradations.}


\bibliographystyle{IEEEtran}
\bibliography{IEEEabrv, refer_paper}

\vfill

\end{document}